\documentclass{article}

\usepackage{PRIMEarxiv}

\usepackage[utf8]{inputenc} 
\usepackage[T1]{fontenc}    

\usepackage{amssymb}
\usepackage{pdflscape}
\usepackage{multirow}
\usepackage{booktabs}
\usepackage{caption}
\usepackage[edges]{forest}
\usepackage{tabularx}
\usepackage{graphicx}
\usepackage[authoryear]{natbib}
\usepackage{enumitem}

\usepackage{hyperref}
\usepackage[draft]{todonotes}
\usepackage{soul}
\sethlcolor{lime}

\usepackage[framemethod=TikZ]{mdframed}
\usepackage{seqsplit}

\usepackage{float}
\usepackage{adjustbox}

\usepackage{tikz}
\usetikzlibrary{shapes, shapes.geometric, matrix, positioning, arrows.meta, fit, calc, backgrounds, shadows, arrows}

\usepackage{longtable}
\usepackage{overpic}

\title{Leveraging AI and NLP for Bank Marketing:\\ A Systematic Review and Gap Analysis
}

\author{
  \href{https://orcid.org/0000-0002-7579-0751}{\includegraphics[scale=0.06]{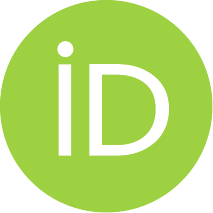}\hspace{1mm}Christopher Gerling} \\
  Chair of Information Systems \\ 
  Humboldt University of Berlin\\  
  Berlin, Germany \\
  \texttt{gerlingc@hu-berlin.de} \\
   \And
 \href{https://orcid.org/0000-0001-7685-262X}{\includegraphics[scale=0.06]{orcid.pdf}\hspace{1mm}Stefan Lessmann} \\
  Chair of Information Systems \\ 
  Humboldt University of Berlin\\  
  Berlin, Germany \\
  \texttt{} \\
}

\begin{document}
\maketitle

\begin{abstract}

This paper explores the growing impact of AI and NLP in bank marketing, highlighting their evolving roles in enhancing marketing strategies, improving customer engagement, and creating value within this sector.
While AI and NLP have been widely studied in general marketing, there is a notable gap in understanding their specific applications and potential within the banking sector. This research addresses this specific gap by providing a systematic review and strategic analysis of AI and NLP applications in bank marketing, focusing on their integration across the customer journey and operational excellence.

Employing the PRISMA methodology, this study systematically reviews existing literature to assess the current landscape of AI and NLP in bank marketing. Additionally, it incorporates semantic mapping using Sentence Transformers and UMAP for strategic gap analysis to identify underexplored areas and opportunities for future research.

The systematic review reveals limited research specifically focused on NLP applications in bank marketing. The strategic gap analysis identifies key areas where NLP can further enhance marketing strategies, including customer-centric applications like acquisition, retention, and personalized engagement, offering valuable insights for both academic research and practical implementation.

This research contributes to the field of bank marketing by mapping the current state of AI and NLP applications and identifying strategic gaps. The findings provide actionable insights for developing NLP-driven growth and innovation frameworks and highlight the role of NLP in improving operational efficiency and regulatory compliance. This work has broader implications for enhancing customer experience, profitability, and innovation in the banking industry.

\end{abstract}


\section{Introduction}
\label{sec:intro}

The integration of Artificial Intelligence (AI) is rapidly reshaping bank marketing strategies, marking a new era in how financial institutions engage clients. This builds upon foundational digital marketing principles in customer relationship management (CRM) and personalization, as highlighted by \cite{Kannan2017}, who emphasized the transformative role of digital technologies in enhancing the marketing mix (Product, Price, Place, Promotion). AI represents a substantial technological shift and offers new opportunities in data processing and predictive analytics. In the banking sector, AI's role is particularly significant due to the industry's unique challenges, including the intangibility of products, higher security requirements, and the need for complex computations within strict regulatory frameworks, necessitating distinct AI-driven marketing approaches.

AI can process unstructured data in order to enhance value and revenue in bank marketing. This potential for deeper insights into consumer behavior, highlighted by \cite{Wedel2016}, remains underexplored, partly due to confidentiality issues in customer data. The use of Natural Language Processing (NLP) as a marketing tool has not been fully leveraged, indicating a gap between AI's potential and its current use in bank marketing. This motivates a two-step approach: first, conducting a systematic review to synthesize existing knowledge, and second, performing a strategic gap analysis using advanced semantic mapping techniques to identify opportunities for NLP-enhanced bank marketing strategies.

This research aims to systematically review and synthesize the literature on AI in bank marketing, with a particular focus on uncovering the potential and opportunities of NLP within this field. Specifically, the goal is to examine how NLP can enhance customer engagement, streamline operations, and improve decision-making. The aim is to identify applications and opportunities for value creation and provide strategic recommendations for effective NLP utilization by bank marketers.

Previous studies have reviewed NLP applications in general marketing (e.g., \cite{Berger2020, Mustak2021, Shankar2022}), examining how automated textual analysis, topic modeling, and sentiment analysis can generate strategic insights. Within finance, NLP reviews primarily address operational efficiency and consumer behavior (e.g., \cite{Kumar2016, Aziz2022}). However, few studies focus on the application of NLP specifically in bank marketing, where unique challenges related to regulatory constraints and complex customer interactions persist. This gap underscores the need for targeted research to explore how NLP can be adapted to the distinct requirements of bank marketing.

This study contributes to the literature on AI in bank marketing as follows: First, using the PRISMA methodology \citep{moher2009}, it systematically reviews (1) AI and advanced analytics applications in bank marketing and (2) NLP-specific implementations in general marketing, identifying current trends.
Second, it combines both PRISMA results in an AI-enhanced gap analysis to highlight underexplored areas and opportunities for NLP-based bank marketing strategies. The gap analysis employs semantic mapping techniques, specifically Sentence Transformers with UMAP, to uncover new research avenues and visualize the landscape of existing studies.
Finally, the analysis shows how NLP can transform customer engagement, marketing strategy, and operational excellence to drive growth and innovation in banking.
\section{Prisma Methodology and Data Collection}
\label{sec:methodology}

This section outlines the methodology for the first step of this study: the systematic review of AI applications in bank marketing, focusing specifically on the potential of NLP.

\subsection{PRISMA Methodology}
This review follows the PRISMA (Preferred Reporting Items for Systematic Reviews and Meta-Analyses) framework to thoroughly explore AI and NLP applications in bank marketing \citep{moher2009}. A detailed search strategy is implemented using Scopus as the primary database for identifying the relevant literature. Scopus was selected because of its extensive coverage of peer-reviewed journals in various disciplines, including finance, technology, and marketing. The search strategy aimed to capture a broad spectrum of AI applications, particularly through digital CRM and personalized marketing.

\begin{figure}[ht!] 
    \centering 
    \includegraphics[width=\linewidth]{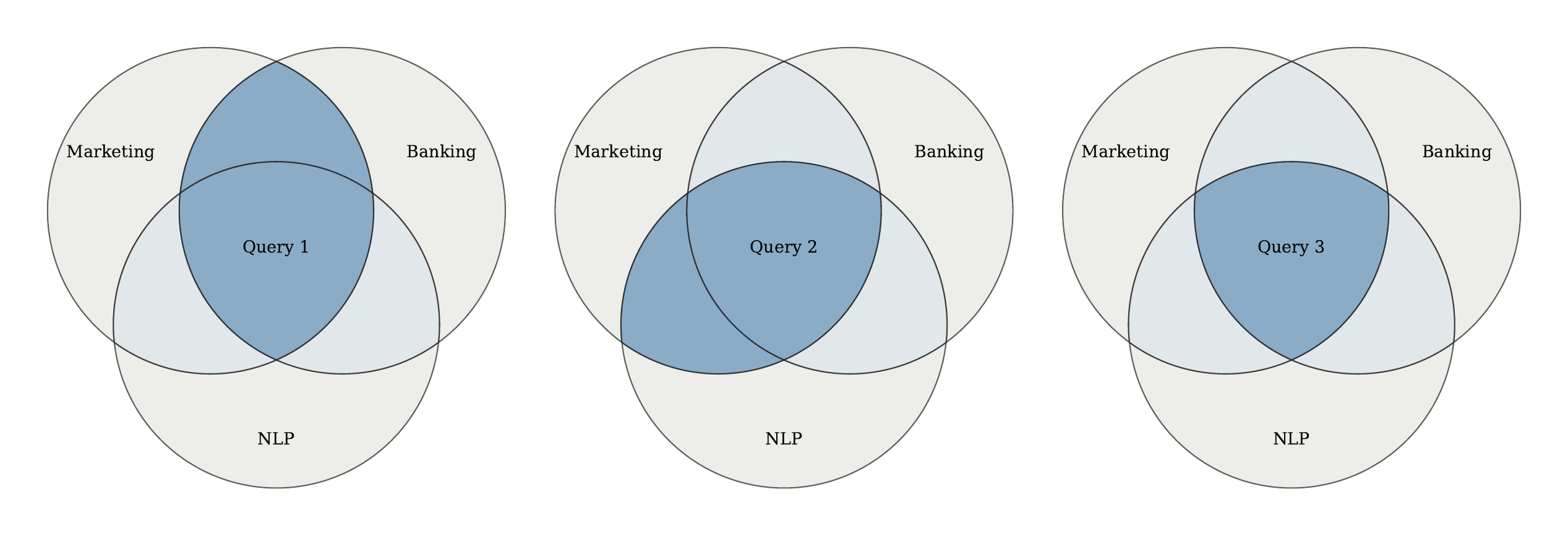} 
    \caption{Areas of interest for PRISMA.} 
    \label{fig:venn-prisma}
\end{figure}

The search strategy is illustrated in Figure \ref{fig:venn-prisma}, using Venn diagrams to depict three primary areas of exploration and their intersections:

\begin{enumerate} 
    \item \textbf{Query 1:} Focuses on analytical marketing in the banking sector (\texttt{MarketingBanking}), providing insights into digital marketing strategies. 
    \item \textbf{Query 2:} Examines marketing and NLP (\texttt{MarketingNLP}), aiming to reveal applications of text processing and generation across industries. 
    \item \textbf{Query 3:} Targets the intersection of marketing, banking, and NLP (\texttt{AllIntersect}), crucial for identifying gaps and opportunities in the literature.  
\end{enumerate}

This scheme provides a conceptual roadmap for identifying and analyzing studies within each intersection. The search was operationalized through specific Boolean search queries detailed in Table \ref{tab:banking_search_queries}. Although tailored to the areas and intersections identified, these queries are kept deliberately broad to capture the maximum range of relevant literature.

\begin{table}[ht!]
\caption{PRISMA Search Queries for AI Applications in Bank Marketing.}
\label{tab:advanced_search_queries}
\centering
\small 
\renewcommand{\arraystretch}{1.2} 
\begin{tabular}{@{}p{0.2\linewidth} p{0.4\linewidth} p{0.3\linewidth}@{}}
\toprule
\textbf{Identifier} & \textbf{Search Query} & \textbf{Description} \\
\midrule
\texttt{MarketingQuery} & "marketing" OR "market analy*" OR "consumer behav*" OR "sales analy*" OR "product recommend*" OR "customer relation*" OR "CRM" OR "pricing" OR "promotion" OR "personali*" OR "churn" & Encompasses broad aspects of analytical marketing, including  customer relationships. \\
\texttt{BankingQuery} & "bank*" OR "financ*" OR "loan"  OR "credit" OR "money" & Covers various facets of banking and financial services. \\
\texttt{NLPQuery} & "NLP" OR "natural language processing" OR "text analy*" OR "sentiment analy*" OR "language model*" OR "speech recog*" OR "chatbot*" OR "LLM" & Focuses on different elements of NLP and its applications. \\
\midrule
\texttt{MarketingBanking*} & \texttt{MarketingQuery} AND \texttt{BankingQuery} & Involves the overlap of analytical marketing in banking. \\
\texttt{MarketingNLP*} & \texttt{MarketingQuery} AND \texttt{NLPQuery} & Highlights the use of NLP in the realm of marketing. \\
\texttt{BankingNLP} & \texttt{BankingQuery} AND \texttt{NLPQuery} & Explores the application of NLP in the banking sector. \\
\midrule
\texttt{AllIntersect**} & \texttt{MarketingQuery} AND \texttt{BankingQuery} AND \texttt{NLPQuery} & Represents the target of NLP-based bank marketing. \\
\bottomrule
\end{tabular}
\begin{flushleft}
\footnotesize{\hspace{1.5em}\textit{*Key queries in the PRISMA framework for this study.}}\\
\footnotesize{\hspace{1.5em}\textit{**Derived from combining individual query outcomes, eliminating a separate search.}}
\end{flushleft}

\label{tab:banking_search_queries}
\end{table}

The selection criteria were designed to ensure high-quality studies. Only peer-reviewed articles published between 2014 and 2024 were included, to reflect advancements in NLP after the introduction of Word2Vec \citep{mikolov2013}. The criteria included journals ranked C or higher, according to the ABDC Journal Quality List, and conference papers were excluded to maintain consistency and quality.

\subsubsection{Data Extraction and Analysis} Data extraction and analysis were conducted in several iterations to ensure thorough coverage and understanding:

\begin{enumerate} \item \textbf{Identification:} Initial database searches using the pre-defined Boolean queries yielded a preliminary pool of records. \item \textbf{Screening:} Records were screened based on titles and keywords, filtering out irrelevant studies and removing duplicates. \item \textbf{Eligibility:} Abstracts were reviewed for eligibility, focusing on analytical bank marketing and NLP-based marketing. Only peer-reviewed journal articles were included. \item \textbf{Inclusion:} The final selection of studies was based on their direct relevance to the research objectives. \item \textbf{Additional Sources:} Additional studies relevant to the research field were included if not identified in the initial search. \end{enumerate}

The iterative process led to a thematic synthesis that identified key themes in this first part of the study. Research gaps will be identified at a later stage.

\section{Findings from PRISMA Analysis}
\label{sec:prisma-results1}

\subsection{Bibliometric and Publication Analysis}
\begin{figure}[h!]
\centering
\begin{tikzpicture}[
    node distance=1cm and 0.4cm,
    auto,
    block/.style={
        rectangle,
        draw,
        text width=2.5cm,
        align=center,
        rounded corners,
        minimum height=1.8cm,
        font=\small
    },
    line/.style={
        draw, -Latex
    },
    header/.style={
        fill={rgb,255:red, 215; green,221; blue, 225},
        fill opacity=1,
        text opacity=1,
        text width=2.5cm,
        align=center,
        rounded corners,
        minimum height=1.8cm
    },
    group/.style={
      rectangle,
      draw,
      dashed,
      inner sep=0.25cm
    }
]

\node[header] (idheader) {Identification};
\node[header, below=of idheader] (scrheader) {Screening};
\node[header, below=of scrheader] (elheader) {Eligibility};
\node[header, below=of elheader] (inchheader) {Included};

\node[block, right=15pt of idheader] (idcontent) {Database records identified (n=7940)};
\node[block, right=15pt of scrheader] (scrcontent) {Title and Keywords screened (n=428)};
\node[block, right=15pt of elheader] (elcontent) {Articles assessed for eligibility (n=84)};
\node[block, right=15pt of inchheader] (inccontent) {Studies included in review (n=35)};

\node[block, right=12pt of scrcontent] (excluded) {Records excluded (n=344) \\ (41 duplicate)};
\node[block, right=12pt of elcontent] (excludedfull) {Articles excluded, with reasons (n=52)};
\node[block, right=12pt of inccontent] (addcontent) {Additionally included (n=3)};

\node[block, right=of idcontent, xshift=3.7cm] (idcontent2) {Database records identified (n=5283)};
\node[block, right=of scrcontent, xshift=3.7cm] (scrcontent2) {Title and Keywords screened (n=680)};
\node[block, right=of elcontent, xshift=3.7cm] (elcontent2) {Articles assessed for eligibility (n=106)};
\node[block, right=of inccontent, xshift=3.7cm] (inccontent2) {Studies included in review (n=82)};

\node[block, right=12pt of scrcontent2] (excluded2) {Records excluded (n=574) \\ (56 duplicate)};
\node[block, right=12pt of elcontent2] (excludedfull2) {Articles excluded, with reasons (n=26)};
\node[block, right=12pt of inccontent2] (addcontent2) {Additionally included (n=2)};

\node[group, fit=(idcontent) (addcontent), label=above:Query 1: \texttt{MarketingBanking}] (crm) {};
\node[group, fit=(idcontent2) (addcontent2), label=above:Query 2: \texttt{MarketingNLP}] (crm) {};

\draw[line] (idcontent) -- (scrcontent);
\draw[line] (scrcontent) -- (elcontent);
\draw[line] (elcontent) -- (inccontent);
\draw[line] (addcontent) -- (inccontent);

\draw[line] (scrcontent) -- (excluded);
\draw[line] (elcontent) -- (excludedfull);

\draw[line] (idcontent2) -- (scrcontent2);
\draw[line] (scrcontent2) -- (elcontent2);
\draw[line] (elcontent2) -- (inccontent2);
\draw[line] (addcontent2) -- (inccontent2);

\draw[line] (scrcontent2) -- (excluded2);
\draw[line] (elcontent2) -- (excludedfull2);
\end{tikzpicture}
\vspace{5px}
\caption{PRISMA Flowchart of Literature Selection.}
\label{fig:prisma_q1_q2}
\end{figure}
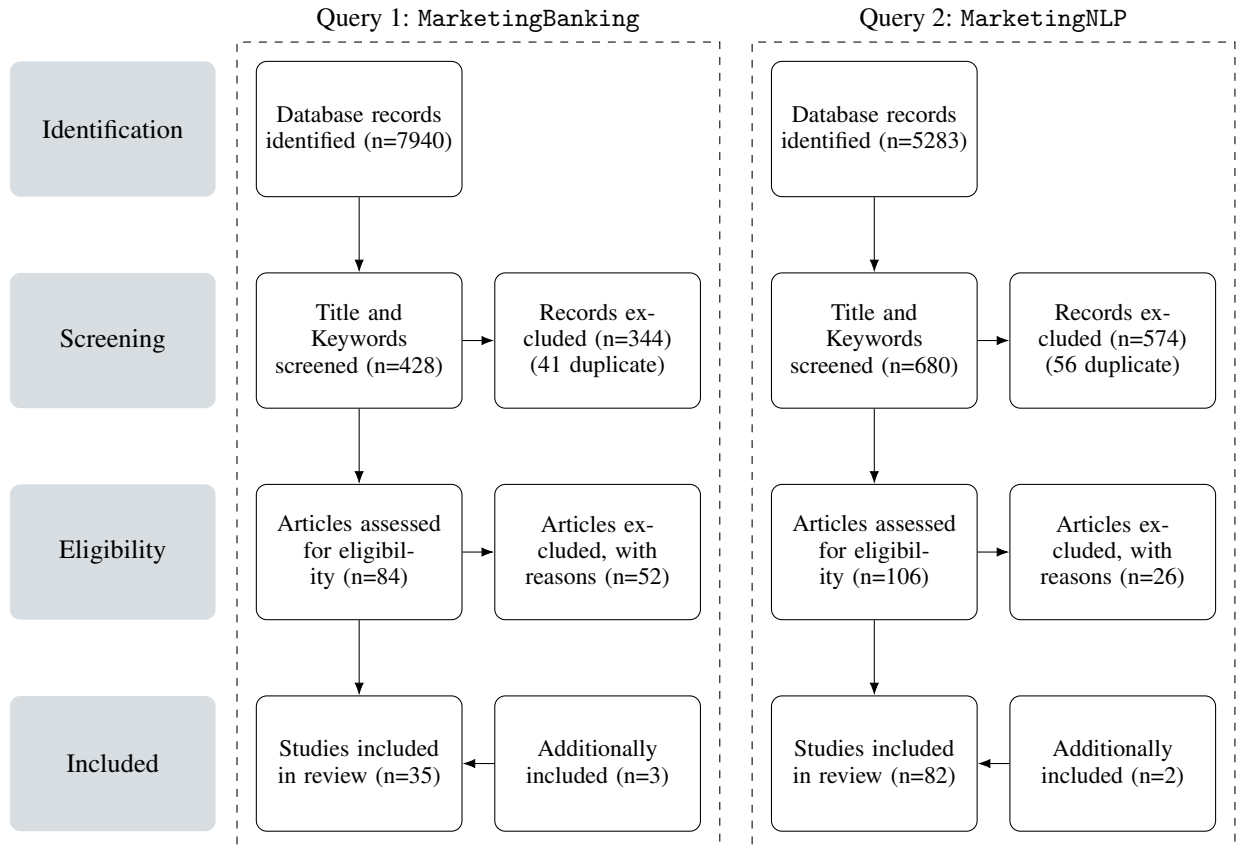

The PRISMA flowchart, shown in Figure \ref{fig:prisma_q1_q2}, visualizes the progression of records through the different phases of the systematic review, specifically for Queries 1 and 2 (\texttt{MarketingBanking} and \texttt{MarketingNLP}). These two queries represent the key areas of focus for this initial step, exploring AI applications in bank marketing and NLP-based marketing, respectively. The third query, \texttt{AllIntersect}, which represents the intersection of marketing, banking, and NLP, yielded a limited number of studies. Therefore, this intersection will be analyzed in a later stage to identify research gaps and potential opportunities as part of the gap analysis.

\begin{figure}[ht!]
    \centering
    \includegraphics[width=\linewidth]{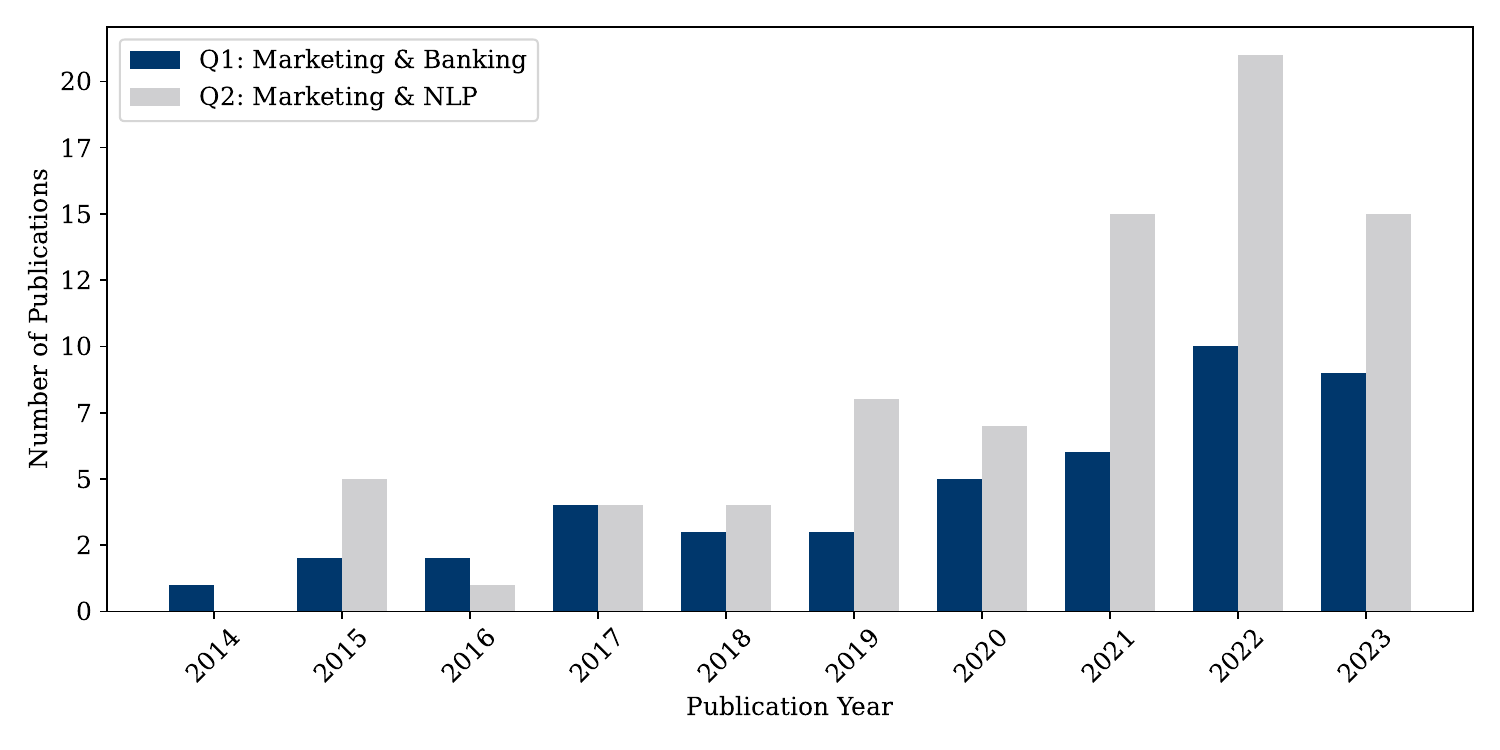} 
    \caption{Number of Publications per Year.}
    \label{fig:prisma_included}
\end{figure}

The publication trend from 2014 to 2023 is illustrated in Figure \ref{fig:prisma_included}, which highlights the yearly frequency of included studies. The data reveals a consistent increase in publications for both \texttt{MarketingBanking} and \texttt{MarketingNLP}, with notable growth in the last years. This trend indicates growing scholarly interest, possibly driven by methodological innovations such as transformer-based language models and evolving industry demands for NLP technologies in marketing.

\newpage
Table \ref{tab:leading_journals} aggregates the journals that currently prevail in the fields of analytical bank marketing and NLP-based general marketing. The publication landscape is more concentrated in specialized journals than in broad multidisciplinary journals.

\begin{table}[htbp]
\centering
\small
\caption{Publication Count in Marketing Research by Journal.}
\label{tab:leading_journals}
\begin{tabular}{p{5cm}p{9.3cm}c}
\toprule
\textbf{Query} & \textbf{Journal} & \textbf{Count} \\
\midrule
\multirow{3}{*}{\texttt{MarketingBanking}} & Expert Systems with Applications & 5 \\
& Decision Support Systems & 2 \\
& Knowledge-Based Systems & 2   \\
& Journal of Business Research & 2 \\

\midrule
\multirow{8}{*}{\texttt{MarketingNLP}} & Journal of Business Research & 7 \\
& Journal of Retailing and Consumer Services & 7 \\
& Technological Forecasting and Social Change & 4 \\
& Journal of Marketing Analytics & 4 \\
& Expert Systems with Applications & 3 \\
& Knowledge-Based Systems & 3 \\
& Electronic Commerce Research & 3 \\
& Journal of Marketing & 3 \\
\bottomrule
\end{tabular}
\end{table}

\subsection{Thematic Classification}

An exploratory analysis of the reviewed literature allowed us to group the two queries into three main dimensions: CRM, marketing mix, and strategic insights. These dimensions are also supported by \cite{Kannan2017} and \cite{Shankar2022}.  This classification provides a structured approach to understanding how AI and NLP enhance different aspects of bank marketing and lays the foundation for a deeper exploration of these themes in the following sections. Table \ref{tab:thematic} presents the thematic distribution of the reviewed studies.

\begin{table}[htbp]
\centering
\small
\caption{Thematic Distribution and Count of Empirical AI/NLP Studies in Marketing.}
\label{tab:thematic}
\noindent
\begin{tabularx}{\textwidth}{Xllrr}
\toprule
Area & Category & Main Focus & \multicolumn{1}{l}{\texttt{MarketingBanking}} & \multicolumn{1}{l}{\texttt{MarketingNLP}} \\
\midrule
Actions & CRM & Customer Acquisition & 2 & 1 \\
& CRM & Customer Retention & 12 & 4 \\
\addlinespace
& Marketing Mix & Product & 1 & 10 \\
& Marketing Mix & Price & 2 & 3 \\
& Marketing Mix & Place & 1 & 0 \\
& Marketing Mix & Promotion & 10 & 20 \\
\addlinespace
Insights & Strategic Insights & Competitor Analysis & 1 & 11 \\
& Strategic Insights & Feedback Analysis & 2 & 49 \\
& Strategic Insights & Other Insights & 10 & 14 \\
\addlinespace
\bottomrule
\end{tabularx}
\end{table}

The \textbf{CRM} dimension highlights a stronger focus on retention over acquisition, underscoring banks' emphasis on long-term relationships. In the \textbf{Marketing Mix}, AI and NLP are primarily applied to product innovation and promotion, while pricing and placement receive less focus. The \textbf{Strategic Insights} dimension captures AI and NLP’s role in competitor and feedback analysis, which supports data-driven decision-making.

In the next subsection, we will classify the selected \texttt{MarketingBanking} papers and analyze them in detail to better understand how each dimension shapes the application of AI and NLP in bank marketing.


\subsection{Analytical Marketing in Banking (Prisma Query 1)}

Building on the thematic classification, this section provides an in-depth analysis of the current research on the application of AI and advanced analytics in bank marketing. It focuses on the three key areas previously identified: CRM, marketing mix, and strategic insights. Within these areas, various research streams are aggregated according to the predefined dimensions, allowing for a detailed exploration of how AI and NLP are being utilized. A detailed summary of these research studies and their key focus areas is depicted in \autoref{tab:bank-mix}.

\newcolumntype{R}[2]{%
    >{\adjustbox{angle=#1,lap=\width-(#2)}\bgroup}%
    l%
    <{\egroup}%
}
\newcommand*\rot{\multicolumn{1}{R{45}{1em}}}
\renewcommand{\arraystretch}{1.3}

\begin{landscape}
\begin{scriptsize}
\begin{longtable}{p{0.08\linewidth}p{0.23\linewidth}p{0.15\linewidth}p{0.15\linewidth}cc@{\hskip 0.5cm}cccc@{\hskip 0.5cm}ccc}
\caption{Chronology of Analytical Bank Marketing.}
\label{tab:bank-mix}\\
Study & Aim & Methods & Data & \rot{Acquisition} & \rot{Retention}&\rot{Product} &  \rot{Price} & \rot{Place} & \rot{Promotion}  & \rot{Competition} & \rot{Feedback} &\rot{Other Insights}  \\
\midrule
\endfirsthead
Study & Aim & Methods & Data & \rot{Acquisition} & \rot{Retention}&\rot{Product} &  \rot{Price} & \rot{Place} & \rot{Promotion}  & \rot{Competition} & \rot{Feedback} &\rot{Other Insights}  \\
\midrule
\endhead
\midrule
\multicolumn{13}{r}{{Continued on next page}} \\
\midrule
\endfoot
\bottomrule
\endlastfoot

\cite{GürAli2014} & To improve customer churn prediction with horizon-specific classifiers & Logistic Regression, Survival Analysis, Cox Regression & Private Banking customer records from a European bank &-&\checkmark  &-&-&-&-  &-&-&-\\

\cite{Abdolvand2015} & To integrate performance management and customer management for strategic decision-making, measuring the Customer Lifetime Value & Genetic K-means, Analytic Hierarchical Processes, Data Envelopment Analysis & Data from an Iranian commercial retail bank &-&-  &-&-&-&-  &-&-&\checkmark\\

\cite{Musto2015} & Improve advisory process in wealth management with individualized asset allocation recommendation framework & Case-based reasoning & Financial Data from an experimental session with 1172 real users  &-&-  &\checkmark&-&-&\checkmark  &-&-&-\\

\cite{Estrella-Ramón2016} & Calculate a multi-product model of Customer Potential Value & Probit method & 24 months of behavioral panel data for 2,187 customers of a Spanish bank &-&-  &-&-&-&-  &-&-&\checkmark\\

\cite{Quijano-Sanchez2017} & To develop a decision support system (DSS) for client acquisition applied to financial networks & Reliability Graph, Probability Function, Projected Gradient Descent, Maximum Reliability Path Algorithm & Client and operations data from Banco Santander, including relationship types, intensity, and expert evaluations  &\checkmark&-  &-&-&-&(\checkmark)  &-&-&-\\

\cite{Schwartz2017} & To develop a DSS for optimizing online display ad impressions to maximize customer acquisition & Multi-Armed Bandit (Reinforcement Learning), Hierarchical Models, Thompson Sampling & Over 750 million ad impressions data from a large retail bank's online display ad campaign  &\checkmark&-  &-&-&\checkmark&(\checkmark)  &-&-&-\\

\cite{Michel2017} & To improve the effectiveness and efficiency of customer selection for direct marketing campaigns using net scoring to forecast campaign-related uplifts & Decision Trees, \(\chi^2\)-statistic & Empirical data from the financial sector involving a dataset with 107,259 customers in the target group  &-&\checkmark  &-&-&-&\checkmark  &-&-&\checkmark\\

\cite{Afolabi2017} & To help the banking industry in Nigeria perform social media competitive analysis and transform social media data into actionable insights & Text mining, Sentiment analysis, K-means clustering & Data from Facebook and Twitter accounts of the five largest banks in Nigeria &-&-  &-&-&(\checkmark)&-  &\checkmark&-&(\checkmark)\\

\cite{Mosavi2018} & To systematically integrate several data mining techniques and management issues to analyze customer value in Tejarat Bank branches in Iran & Fuzzy Analytic Hierarchy Process (FAHP), K-means clustering, Decision tree, Support Vector Machine (SVM), Random forest & Demographic, frequency, money, and trust attributes of customers from Tejarat Bank branches in Iran  &-&-  &-&-&-&-  &-&-&\checkmark\\

\cite{Chakrabarti2018} & Analyzing user reviews to better understand the correlation between RATER dimension sentiment scores and user overall rating (customer satisfaction)& Logistic Regression, Sentiment Analysis & Online reviews from BankBazaar.com for three largest private banks in India  &-&-  &-&-&-&-  &-&\checkmark&(\checkmark)\\

\cite{Shirazi2019} & Constructing a predictive churn model using big data and analyze various aspects of customer behavior affecting churn in the Canadian banking industry&  Classification and Regression Trees (CRT) & Structured archival data and unstructured data (online web pages, website visits, phone conversation logs) from over 3 million customer records  &-&\checkmark  &-&-&-&-  &-&-&-\\

\cite{Bogaert2019} &Evaluate multi-label classification techniques and recommender systems for cross-sell purposes in the financial services sector& Multi-label classification techniques (binary relevance, classifier chains with base classifiers such as Adaboost, random forest, neural networks), Recommender systems (user-based collaborative filtering, item-based collaborative filtering, association rules, popular recommender method, random recommender method) & Customer data from an international financial services provider, including 2,923,989 unique customers and 6808 unique products &-&-  &-&-&-&\checkmark  &-&-&-\\

\cite{Ladyzynski2019} & To develop a system for predicting customers' willingness to take a personal loan using deep learning and random forests. & Random forests, Deep belief networks & Transactional and historical data from a large retail bank in Poland  &-&-  &-&-&-&\checkmark  &-&-&-\\

\cite{Coşer2020} & Identifying key characteristics and predicting the likelihood of customer churn in banking & Logistic Regression, Random Forest (with Grid Search optimization) & Bank customer data from Kaggle &-&\checkmark  &-&-&-&-  &-&-&-\\

\cite{DeCaigny2020} & Improve customer churn prediction by incorporating textual data from customer communications & Logistic Regression, CNN & Financial services provider data, text messages  &-&\checkmark  &-&-&-&-  &-&-&-\\

\cite{Vélez2020} & Forecasting customer churn and Net Promoter Score (NPS) to improve decision-making and customer retention & Logistic Regression, Stepwise Selection, WOE Variables, Gradient Boosting & Client survey data from a global bank, client database records  &-&\checkmark  &-&-&-&-  &-&-&-\\

\cite{Hernández-Nieves2020} &Developing a fog computing-based architecture to improve personalized customer service through predictive systems for recommending banking products& Fog computing architecture, Hybrid recommendation system, CBR & Data from mobile devices, applications, and local bank equipment processed through fog nodes with intelligent agents  &-&-  &-&-&-&\checkmark  &-&-&-\\

\cite{Mancisidor2021} & Improve marketing campaigns, customer relationship management, data and process management, and credit risk assessment by using VAE to learn latent representations that reflect customers' creditworthiness & Variational Autoencoder (VAE), AEVB algorithm, Weight of Evidence (WoE) transformation & Norwegian and Finnish car loan data, Kaggle "Give Me Some Credit" dataset  &-&-  &-&-&-&-  &-&-&\checkmark\\

\cite{Vo2021} & Develop a customer churn prediction model leveraging unstructured data from customer call logs to enhance churn risk prediction accuracy and provide retention insights & Text Mining with TF-IDF, Multi-stacking ensemble (XGBoost, Logistic Regression, Random Forest, Gaussian Naive Bayes) & Two million call logs from over two hundred thousand customers, provided by an Australian financial services company  &-&\checkmark  &-&-&-&-  &-&-&(\checkmark)\\

\cite{Piris2021} & To analyze customer feedback and identify key factors influencing customer satisfaction & Latent Dirichlet Allocation (LDA), Data pre-processing, Bag-of-Words (BoW), Term Frequency-Inverse Document Frequency (TF-IDF) & 12,000 Customer Feedback Answers from a Bank Survey &-&-  &-&-&-&-  &-&\checkmark&-\\

\cite{Ho2021} & Examine the longitudinal relationships between customer loyalty behaviors and firm financial outcomes in the context of credit card usage & Sequence Analysis, Optimal Matching, Cluster Analysis, Activity-Based Costing (ABC) & Longitudinal dataset from a national bank in Taiwan, including 87,667 credit card customers  &-&\checkmark  &-&-&-&-  &-&-&-\\

\cite{Ban2021} & Develop a personalized dynamic pricing model using high-dimensional customer features to optimize pricing and maximize revenue & Personalized demand model, Semi-clairvoyant policy design, Near-optimal pricing policy, Lasso regularization, Maximum quasi-likelihood regression & Simulated data, Real-life data from an online auto loan company in the United States &-&-  &-&\checkmark&-&-  &-&-&(\checkmark)\\

\cite{Shumanov2022} & Using AI predicted personality to enhance advertising effectiveness & Mixed-method approach, AI algorithms (GloVe, machine learning) & Retail banking customer data, survey responses, voice-to-text customer interaction data  &-&-  &-&-&-&\checkmark  &-&-&\checkmark\\

\cite{AL-Najjar2022} & Develop credit card customer churn prediction models using ML to generate early alerts for banks, enabling proactive customer retention strategies & Bayesian network, C5 tree, CHAID tree, CR tree, neural network, k-nearest neighbor clustering & Credit card customer data from Analyttica, 10,127 customers, 20 variables &-&\checkmark  &-&-&-&-  &-&-&-\\

\cite{Kim2022} & Validate stochastic customer base models in noncontract settings for financial services & Pareto/NBD, Pareto/GGG, BG/CNBD-k, MBG/CNBD-k & Transaction data from a nationwide financial services company in Korea (2015-2018) &-&\checkmark  &-&-&-&-  &-&-&(\checkmark)\\

\cite{Valluri2022} & Evaluate the effectiveness of ML techniques in predicting used auto loan churn among subprime borrowers & Logistic Regression, Linear Discriminant Analysis, Decision Trees, Random Forests & Dataset from a Midwestern credit union (2014-2016) &-&\checkmark  &-&-&-&-  &-&-&(\checkmark)\\

\cite{Feng2022} & Proposing a dynamic ensemble selection method for predicting bank telemarketing sales & META-DES-AAP (Dynamic Ensemble Selection, Multi-objective Optimization) & Telemarketing dataset from a retail bank in Portugal &-&-  &-&-&-&\checkmark  &-&-&\checkmark\\

\cite{Tékouabou2022} & Proposing a machine learning framework using the CMB approach to predict the success of bank telemarketing campaigns by addressing data heterogeneity and feature preprocessing & Class Membership-Based (CMB) Classifier & UCI Portuguese bank telemarketing dataset (45,000 customers contacted between 2008 and 2012)  &-&-  &-&-&(\checkmark)&\checkmark  &-&-&\checkmark\\

\cite{Rouhani2023} & Developing a hybrid model to predict customer churn in the banking industry & Decision Tree (DT), Multinomial Regression (MR), Hybrid DT-MR & Bank dataset (6507 customers, 10 months) &-&\checkmark  &-&-&-&-  &-&-&-\\

\cite{Radovanović2023} & Proposing a fair classifier chain model for multi-label classification in bank marketing strategies to address fairness issues in predictive modeling & Fair Classifier Chain using Logistic Regression & Historical data from bank marketing campaigns (2240 customers)  &-&-  &-&-&-&\checkmark  &-&-&\checkmark\\

\cite{Simsek2023} & Classifying and predicting customer churn in the banking sector, with a focus on addressing data imbalance for improved prediction accuracy. & Logistic Regression, K-Nearest Neighbor, Support Vector Machine, Decision Tree, Random Forest, Gradient Boosting Machines, LightGBM, XGBoost, ROS-Voting (RF-LGBM) & Kaggle dataset of churning bank customers &-&\checkmark  &-&-&-&-  &-&-&\checkmark\\

\cite{Natesan2023} & Formulating dynamic and static pricing models for home loans to optimize net present value over 15 years using various demand functions and default probability relationships & AMPL with MINOS solver, quadratic programming, sensitivity analysis & Data from a nationalized bank in India (demand for home loans over different interest rates) &-&-  &-&\checkmark&-&-  &-&-&-\\

\cite{Boustani2023} & Improving the predictive accuracy of cross-selling consumer loans in retail banking by utilizing deep learning networks and account transaction data. & Deep Learning (Auto-encoder, Neural Networks, Decision Trees, SVM, Random Forest, Gradient Boosting Machine, Heterogeneous Ensemble) & Transaction data of 6127 individual customers from a Middle Eastern bank, including demographics, product ownership, and credit card transactions &-&-  &-&-&-&\checkmark  &-&-&-\\

\end{longtable}
\end{scriptsize}
\end{landscape}

\subsubsection{Customer Relationship in AI-driven Bank Marketing}
The enhancement of customer relationships through AI technologies exemplifies a multi-dimensional approach. Specifically, AI has been instrumental in transforming how banks acquire and retain customers.

\paragraph{Customer Acquisition.}
AI-driven customer acquisition strategies have demonstrated significant potential for optimizing marketing efforts. For instance, \cite{Quijano-Sanchez2017} developed a decision support system for Banco Santander, which utilized client and operational data to identify the most reliable sequences for client acquisition and cross-selling. This decision support system employed state-of-the-art models, including a reliability graph and a probability function, to optimize the client acquisition process. Similarly, \cite{Schwartz2017} optimized online display ad impressions using multi-armed bandit methods and hierarchical models, achieving an 8\% improvement in customer acquisition without additional costs. Adaptive learning and real-time data optimization are thus effective methods to enhance customer acquisition strategies in banks.

\paragraph{Customer Retention.}
Predictive models for customer retention have been instrumental in reducing churn rates, enabling banks to maintain a stable customer base and mitigate revenue losses. Various studies have employed advanced AI and ML techniques to enhance these predictive models and develop effective retention strategies.

\cite{GürAli2014} introduced a dynamic churn prediction framework using horizon-specific binary classifiers. Their Multiple Period Training Data (MPTD) approach increased the size of the training data set and reduced the rarity of churn events, thereby enhancing prediction accuracy. The study makes use of logistic regression, survival analysis, and Cox regression on customer records from a European bank and specifically deals with rate churn events.

Leveraging big data analytics, \cite{Shirazi2019} developed a model to identify churn patterns among retirees in the Canadian banking industry. By analyzing unstructured data from online web pages, website visits, and phone conversation logs, the study provided a view of customer behavior, enabling targeted retention strategies. The integration of diverse data sources has been a subject of investigation in the field of churn prediction modeling. \cite{Coşer2020} used a dataset from Kaggle consisting of 10,000 bank customer records with variables such as socio-demographic characteristics, balance, estimated salary, and tenure. They forecast customer attrition using logistic regression and random forest. The random forest classifier demonstrated superior predictive accuracy, enabling banks to implement preemptive measures to retain valuable customers. \cite{DeCaigny2020} enhanced churn prediction by incorporating textual data from customer communications with advisors. The study improved the accuracy of churn prediction models by using logistic regression and convolutional neural networks (CNNs), which provided deeper insights into customer behavior and communication patterns.

\cite{Vélez2020} combined logistic regression, stepwise selection, and Weight of Evidence (WOE) variables to forecast customer churn and Net Promoter Score (NPS). The study identified strong predictors, such as customer age and the number of insurance contracts, which enabled targeted actions to retain customers.  Similarly, \cite{Vo2021} leveraged unstructured call log data and text mining techniques, integrating features into a multi-stacking ensemble model. This approach provided deeper insights into customer behavior and risk factors, enabling more effective and personalized retention strategies.

Studies such as that of \cite{Ho2021} have focused on segmenting customers based on their loyalty behaviors to develop targeted retention strategies. Employing sequence analysis, the research segments customers into six distinct groups: Loyalist, Switching Loyalist, Switching Defector, Defector, Dormant Loyalist, and Dormant Defector. These groups show varying profitability, allowing banks to allocate marketing resources more effectively. Tree-based models, exemplified in the work of \cite{AL-Najjar2022} utilizing the C5 decision tree algorithm, are robust predictors of customer churn. Their model identifies key predictors, such as total transaction count and revolving balance, which can be used to inform proactive measures to retain customers and enhance stability.

Moreover, the efficacy of stochastic models in forecasting customer activity and retention in noncontractual settings was validated by \cite{Kim2022} through the use of transaction data to classify active customers with high accuracy. \cite{Valluri2022} evaluated the effectiveness of various ML techniques in predicting used auto loan churn among subprime borrowers, identifying significant variables such as loan-to-value ratio and FICO credit score to inform the development of targeted interventions.

Finally, hybrid models have demonstrated considerable potential for enhancing the accuracy of churn prediction. \cite{Rouhani2023} combined decision trees and multinomial regression to categorize customers into three distinct classes: non-churn, prone to churn, and churn. This approach yielded an average accuracy of 87.66\%. In addressing data imbalance in churn prediction, \cite{Simsek2023} employed a hybrid model combining random oversampling (ROS) and voting (RF-LGBM), achieving high accuracy (95\%) and F1-score (0.95) and effectively identifying at-risk customers. 

Together, these studies collectively underscore the importance of leveraging advanced analytics and AI techniques to enhance customer retention strategies in the banking sector. By accurately predicting customer churn and understanding the underlying factors, banks can develop targeted interventions that maintain a stable customer base and improve profitability.


\subsubsection{Marketing Mix in AI-driven Bank Marketing}
AI technologies enable banks to optimize the 4 P's of the marketing mix, enhancing customer engagement, improving operational efficiency, and driving revenue growth. This section explores how AI-driven approaches have the potential to transform products, pricing strategies, channel selection, and promotional activities.

\paragraph{Product Innovation.}
\cite{Musto2015} explored the enhancement of financial advisory services in wealth management through individualized asset allocation recommendations. To this end, they employed case-based reasoning and diversification strategies, with the objective of creating personalized investment portfolios. The results of their investigation demonstrated that these AI-driven portfolios could align with clients' risk profiles and potentially offer superior yields compared to traditional human advisor recommendations. Consequently, they concluded that the use of AI in wealth management could improve the advisory process.

\paragraph{Pricing Strategies.}
In their respective studies, \cite{Ban2021} and \cite{Natesan2023} explored dynamic pricing models with the objective of optimizing revenue. \cite{Ban2021} demonstrated the effectiveness of personalized dynamic pricing using high-dimensional customer features and heterogeneous price elasticity. The study employed machine learning (ML) techniques such as lasso regularization and maximum quasi-likelihood regression, achieving a 47\% increase in expected revenue over a six-month period. \cite{Natesan2023} formulated dynamic and static pricing models for home loans to optimize the net present value (NPV) of cash flows over a 15-year period. The study used data from a nationalized bank in India, employing quadratic programming. The dynamic pricing model generally yielded higher expected revenue compared to the static model, particularly for linear demand functions. This suggested that continuously adjusting loan interest rates in response to market conditions could significantly enhance a bank's revenue.

\paragraph{Place Strategies.}
\cite{Schwartz2017} addressed optimal channel selection by optimizing online display ad impressions to improve customer acquisition rates. Using multi-armed bandit (MAB) methods and hierarchical models, the study dynamically adjusted ad placements based on real-time data, addressing the "learn-and-earn" trade-off. Implemented in a live field experiment, the system achieved an 8\% improvement in customer acquisition without additional costs, highlighting its effectiveness in enhancing customer acquisition strategies through adaptive learning and real-time data optimization.

\paragraph{Promotional Activities.}
The application of AI has significantly enhanced the efficacy of promotional strategies of banks. AI-driven marketing efforts are now more personalized and efficient than traditional methods. For instance, \cite{Musto2015} demonstrated how AI can be leveraged to enhance wealth management. This was achieved by utilizing AI for individualized investment recommendations through case-based reasoning and diversification strategies. These approaches yielded superior yields and more closely aligned with clients' risk profiles than traditional methods.

\cite{Michel2017} optimized direct marketing campaigns by using decision trees and a new \(\chi^2\)-statistic to accurately forecast campaign-related uplifts, identifying customers likely to respond positively. This approach maximized campaign impact and efficiency. \cite{Bogaert2019} compared multi-label classification techniques and recommender systems. Their findings indicated that user-based collaborative filtering and classifier chains with Adaboost were the most effective for cross-selling in the financial sector. These methods enhanced the prediction and recommendation of pertinent products to customers.

In retail banking, \cite{Ladyzynski2019} utilized ML models, including random forests and deep belief networks, to predict customers' likelihood of purchasing credit products. This approach allowed for optimal timing and personalized offers, increasing the effectiveness of direct marketing campaigns. \cite{Hernández-Nieves2020} integrated fog computing and hybrid recommendation methodologies to provide real-time, personalized product suggestions, such as mortgages and loans. This innovative system enhanced customer satisfaction and conversion rates by ensuring the delivery of timely and relevant product recommendations.

\cite{Shumanov2022} enhanced advertising effectiveness by predicting consumer personality traits and matching them with congruent advertising messages. The study employed AI techniques, including Global Vectors for Word Representation (GloVe) and other ML algorithms, to infer personality traits from voice-to-text customer interaction data. This approach provides a scalable method for enhancing marketing communications and validating AI for accurate personality profiling. The results indicate that extroverted consumers demonstrated a greater responsiveness to advertisements that evoked excitement and social rewards, whereas neurotic consumers exhibited a preference for messages that reduced perceived risks and provided social acceptance cues. This tailored advertising approach led to a notable improvement in click-through and conversion rates.

\cite{Feng2022} aimed to enhance the efficacy of bank telemarketing sales through the implementation of a dynamic ensemble selection method (META-DES-AAP). This method considered both the accuracy and the average profit of the telemarketing campaigns. The insights gained from this method enabled more informed and strategic marketing decisions. \cite{Tékouabou2022} employed a Class Membership-Based (CMB) classifier to improve the prediction of successful telemarketing campaigns. The CMB approach, which processes heterogeneous data such as age, marital status, and educational level, achieved high accuracy (97.3\%) and an AUC score (95.9\%) through precise targeting of potential customers.

The ethical considerations of promotional activities were addressed by \cite{Radovanović2023}, who introduced a fair classifier chain approach to avoid gender bias in predictive models. The study highlighted a trade-off between increased fairness (by 7\% to 17\%) and a slight decrease in predictive performance (up to 9\% in AUC), emphasizing the importance of ethical considerations in marketing strategies.

Finally, \cite{Boustani2023} demonstrated the value of transaction data in improving cross-selling models using deep learning networks. By analyzing nearly 800,000 credit card transactions, the study showed that integrating transaction data with demographic and product ownership data significantly enhanced predictive accuracy. The heterogeneous ensemble model achieved an AUC of 0.932, highlighting the potential for optimized promotional efforts in retail banking.


\subsubsection{Strategic Insights in AI-driven Bank Marketing}
With AI, banks extend marketing beyond traditional metrics, offering insights into competitors and customers. This knowledge streamlines decision-making for market positioning.

\paragraph{Competitor Analysis.}
\cite{Afolabi2017} demonstrated the use of social media data for competitive analysis in Nigeria's banking industry. By employing text mining, sentiment analysis, and k-means clustering techniques, the study analyzed unstructured text content from Facebook and X/Twitter accounts of the largest banks. This analysis provided deep insights into customer sentiments and engagement, enabling banks to understand their competitive positioning and devise strategic marketing initiatives. The study demonstrates the potential for banks to make informed decisions and enhance their competitive edge in the market by transforming social media data into actionable knowledge.

\paragraph{Feedback Analysis.}
In order to gain knowledge about the service quality and satisfaction of customers, the study conducted by \cite{Chakrabarti2018} analyzed online customer reviews. The analysis was conducted using sentiment analysis, which enabled the researchers to identify the dimensions of service that had the greatest impact on customer satisfaction. These dimensions included responsiveness and tangibles. Similarly, \cite{Piris2021} utilized topic modeling to analyze customer feedback from a bank's satisfaction survey, identifying key topics that influence satisfaction levels.

\paragraph{Other Insights.}
Beyond competitor and feedback analysis, AI-driven analytics generate valuable insights that enhance customer relationships and profitability in banking. \cite{Abdolvand2015} proposed integrating performance management with customer management using Customer Lifetime Value (CLV) as a financial metric. This integrative approach employed a variety of techniques, including genetic K-means, analytic hierarchical processes, and data envelopment analysis, to segment customers into distinct clusters. \cite{Estrella-Ramón2016} calculated a Customer Potential Value (CPV) through a probit model and panel data of Spanish banks. The researchers identified a number of key determinants of future profits, including the length of the relationship, product usage, and cross-buying behavior. Furthermore, \cite{Mosavi2018} utilized data mining techniques, including fuzzy analytic hierarchy process (FAHP) and K-means clustering, to construct a customer value pyramid. This approach yielded valuable insights into customer behavior and preferences, thereby facilitating the development of more effective customer relationship management strategies. Additionally, \cite{Mancisidor2021} employed a Variational Autoencoder (VAE) to learn latent representations of bank customers' data, specifically to reflect their creditworthiness. By transforming the input data using Weight of Evidence (WoE), the VAE created meaningful latent spaces that clustered customers based on their likelihood of default. This clustering allowed banks to better understand customer segments and to enable a more efficient customer relationship management. These customer insights are frequently linked with subsequent marketing actions for profit generation.

\subsection{Natural Language Processing in Marketing (Prisma Query 2)}
\label{sec:prisma-results2}

Following the analysis of AI applications in bank marketing, we now turn to the broader landscape of general marketing studies utilizing NLP (\texttt{MarketingNLP}). As a subset of AI, NLP has been widely used to analyze and interpret human language in various marketing contexts. Inspired by \cite{Berger2020}, we begin by examining the \textit{Textual Universe in Marketing}, focusing on text producers and receivers, including sources like social media posts, customer reviews, and call logs. Understanding these data flows is crucial for effective NLP applications.
Next, we explore NLP models and techniques for \textit{Decoding Texts in Marketing}, such as sentiment analysis and topic modeling, which are essential for generating insights. After the publication of \cite{Berger2020}, transformer-based models have further revolutionized the field of NLP, adding new capabilities not covered in their analysis.
Finally, we categorize the reviewed studies by industry sectors to identify potential improvements in banking from other sectors. This second PRISMA review provides the basis for our subsequent gap analysis, which will delve into further applications of NLP in bank marketing.

\subsubsection{Textual Universe in Marketing}
To effectively apply NLP in marketing, it is crucial to understand the sources of textual data. Following \cite{Berger2020}, we distinguish between text producers and receivers, focusing primarily on firms and customers as the key stakeholders.

Table \ref{tab:text_receiver_producer} categorizes the types of texts produced and received by firms and customers, based on the textual data used in the included studies. It illustrates the diversity of textual data available for NLP in analytical marketing. As shown, a substantial portion of the text generated in marketing originates from customers, such as feedback, reviews, social media posts, and user-generated content. These sources are heavily researched for understanding customer sentiment and behavior. Firms, as the primary recipients, derive actionable insights and business value from these data sources, although other customers may also consume the content.

Firms also produce text, with a more balanced distribution between customers and internal recipients. Direct communication with customers, such as marketing messages, online ads, chatbot interactions, and content marketing tweets, has emerged as a significant area of interest. Internally, firms generate reports and operational texts.

\renewcommand{\arraystretch}{1.2}

\begin{landscape}

\begin{table}[h]
    \centering
    \scriptsize
    \caption{Types of Texts Produced and Received by Firms and Customers.}
    \begin{tabular}{@{}lp{0.4\linewidth}p{0.4\linewidth}@{}}
        \toprule
        \textbf{} & \multicolumn{2}{c}{\textbf{Text Receiver}} \\ \cmidrule(lr){2-3}
        \textbf{Text Producer} & \textbf{Firm} & \textbf{Customer} \\ \midrule

        \textbf{Firm} & 
        \begin{minipage}[t]{9cm}
        \textbf{Reports and Logs}:
        \begin{itemize}[leftmargin=*,itemsep=0pt,parsep=3pt,topsep=2pt,partopsep=0pt]
            \item Annual Reports and Financial Filings \citep{Cooper2022, Han2021}
            \item Customer Call Notes and Interaction Logs \citep{Ozan2021, Shumanov2022}
        \end{itemize}
        \textbf{Internal Operations}:
        \begin{itemize}[leftmargin=*,itemsep=0pt,parsep=3pt,topsep=2pt,partopsep=0pt]
            \item Marketing Prompt Templates and Brand Communications \citep{Tafesse2024}

        \end{itemize}
        \end{minipage} & 
        \begin{minipage}[t]{9cm}
        \textbf{Customer Communications}:
        \begin{itemize}[leftmargin=*,itemsep=0pt,parsep=3pt,topsep=2pt,partopsep=0pt]
            \item Marketing Messages and Online Ads \citep{Atalay2023, Wu2022}
            \item Chatbot and Voice Agent Interactions \citep{Kushwaha2021, Rhee2020}
            \item Content Marketing Tweets \citep{Barbosa2023, Batta2023, Galiano-Coronil2023, Liu2021, Gruss2020, Hu2019}
            \item Email Campaign and Advertising Materials \citep{Atalay2023, Wu2022}
            \item Newsletters and Social Media Posts
        \end{itemize}
        \textbf{Marketing Content}:
        \begin{itemize}[leftmargin=*,itemsep=0pt,parsep=3pt,topsep=2pt,partopsep=0pt]
            \item SEO Texts for Website Landing Pages \citep{Reisenbichler2022}
        \end{itemize}
        \end{minipage} \\ \midrule

        \textbf{Customer} & 
        \begin{minipage}[t]{9cm}
        \textbf{Feedback and Reviews}:
        \begin{itemize}[leftmargin=*,itemsep=0pt,parsep=3pt,topsep=2pt,partopsep=0pt]
            \item Customer Reviews from Various Platforms \citep{Marandi2024, Le2024, Nicolau2024, Liu2024, Zhai2024, Park2023, Karn2023, Barbosa2023, Han2023, Solairaj2023, Batta2023, Khan2023, Liu2023, Yazıcı2023, Loupos2023, Rahimi2022, Sun2022, Aldunate2022, GuhaMajumder2022, Kim2022, Garner2022, Alzate2022, Alantari2022, Zhang2022, Farzadnia2022, Vassilikopoulou2022, Celuch2021, Karthik2021, Shambour2021, HeidaryDahooie2021, Zhang2021, Ghasemi2021, Situmeang2020, Deng2020, Timoshenko2019, Lawani2019, Kim2019, Zhang2018, Dubey2017, Ma2017, Lee2016, Xiang2015, Li2015, Yang2015}
        \end{itemize}
        \textbf{User-Generated Content}:
        \begin{itemize}[leftmargin=*,itemsep=0pt,parsep=3pt,topsep=2pt,partopsep=0pt]
            \item Social Media Posts and Tweets \citep{Lappeman2022, Swaminathan2022, Agrawal2022, Lamrhari2022, Wu2022, VillarroelOrdenes2021, Fresneda2021, Gozuacik2021, Kushwaha2021, Piris2021, Lopez2020, Missaoui2019, Cu2019, Chang2019, Hu2019, Ghosh2018, Peláez2018, Afolabi2017, Pathak2017, Balusamy2015}
        \end{itemize}
        \textbf{Direct User Interaction}:
        \begin{itemize}[leftmargin=*,itemsep=0pt,parsep=3pt,topsep=2pt,partopsep=0pt]
            \item Electronic Messages to Client Advisors \citep{DeCaigny2020}
            \item Speech and Facial Expressions (video data) \citep{Chen2022}
        \end{itemize}
        \end{minipage} & 
        \begin{minipage}[t]{9cm}
        \textbf{Social Media Interactions}:
        \begin{itemize}[leftmargin=*,itemsep=0pt,parsep=3pt,topsep=2pt,partopsep=0pt]
            \item Forum Posts and Discussions \citep{Sun2022, Wu2020, Liu2019, Homburg2015}
            \item Blog Posts
            \item Social Media Reviews
        \end{itemize}

        \end{minipage} \\ \bottomrule
    \end{tabular}
    
    \label{tab:text_receiver_producer}
\end{table}

\end{landscape}

\subsubsection{Decoding Texts in Marketing with NLP}
NLP has revolutionized marketing by providing tools to analyze and interpret vast amounts of textual data. This enables marketers to gain valuable insights into customer behavior, preferences, and sentiments. In this section, we explore advanced NLP methodologies and their applications in marketing, as summarized in \autoref{tab:nlp_methods}.

\paragraph{Sentiment and Emotion Analysis.}

A predominant area of interest within NLP studies in marketing is the investigation of customer sentiment and emotions. Sentiment analysis is a technique that identifies and categorizes opinions expressed in texts to determine the overall attitude of an individual or a group towards a particular product or brand. This attitude can be represented numerically or classified into opinion polarity classes, which include positive, neutral, and negative \citep{Barbosa2023, Solairaj2023, Zhang2022, Lappeman2022, GuhaMajumder2022, Afolabi2017}. Such analysis is beneficial for providing timely feedback, managing brand reputation, shaping marketing strategies, and gaining more profound insights into brand perception.

Aspect-Based Sentiment Analysis (ABSA) provides a more granular perspective compared to general sentiment analysis, allowing firms to identify specific features or aspects of products and services that drive customer satisfaction \citep{Liu2024, Lappeman2022, Vassilikopoulou2022}. This technique differentiates between different aspects of products mentioned within the text, providing a targeted understanding of customer feedback.

An even more complex method for analyzing customer sentiments is Multimodal Emotion Recognition \citep{Chen2022}. This method interprets emotions through multiple data types, including speech-to-text and image data. This represents the cutting-edge of sentiment analysis, offering deeper insights that can be used to tailor marketing strategies effectively.

Sentiment Analysis is often integrated with other NLP methods, such as topic modeling, to enhance the understanding of customer feedback \citep{Barbosa2023}. 

\paragraph{Topic Modeling.}
Topic modeling is a set of techniques that can be used to identify hidden topics or trends within textual datasets. In the field of marketing, topic modeling is employed to derive valuable insights about customers, including customer preferences and market trends.

The majority of studies in this field employ Latent Dirichlet Allocation (LDA) with a fixed number $k$ of latent topics to be identified in a text corpus. LDA, for instance, has been effectively used by \cite{Le2024} and \cite{Barbosa2023} to aid in market segmentation and trend analysis.
For example, LDA has been applied to customer reviews and social media posts to uncover underlying themes, enabling businesses to tailor their marketing strategies accordingly \citep{Nicolau2024}.

Closely related to this is Correlated Topic Modeling (CTM), which allows for correlations between topics, providing a nuanced understanding of thematic relationships within a dataset \citep{Garner2022}. This approach is particularly valuable for datasets where topics are likely to be interrelated, such as customer reviews spanning multiple product categories.

Newer methods, such as transformer-based language models like BERT (Bidirectional Encoder Representations from Transformers), are gaining popularity in the field of topic modeling and sentiment analysis \citep{Yazıcı2023}.  By leveraging pre-trained embeddings, these models facilitate more accurate topic identification, thereby improving the semantic understanding of text.

\paragraph{Natural Language Generation.}
The fields of Natural Language Generation (NLG), Conversational AI, and Generative AI (GenAI) have recently attracted significant interest from researchers and practitioners alike, with new applications in marketing emerging.

The rise of Large Language Models (LLMs) such as GPT has facilitated the creation of marketing content through user prompts, offering a new level of versatility in text generation  \citep{Tafesse2024}. In marketing practice, the underlying data and the desired outcome shape the complexity of the prompt engineering process. LLMs enable marketers to create highly personalized content, enhancing customer engagement.

A related development is the integration of LLMs into customer-facing interactions, such as chatbots \citep{Jin2023, Kushwaha2021} or voice systems \citep{Rhee2020}. LLMs can provide the textual basis and analytics tools out of the box to handle customer queries and automate sales processes. These chatbots can simulate human-like interactions, thereby improving customer satisfaction and engagement.

In addition to customer interactions, NLG can also generate high-quality SEO content, which is essential for improving search engine rankings and attracting more visitors to websites. The use of transfer learning in this context can further fine-tune models for specific domains, ensuring the generated content closely mirrors human-written text \citep{Reisenbichler2022}.

\renewcommand{\arraystretch}{1.15}
\begin{table}[H]
    \centering
    \scriptsize
    \caption{NLP Methods in Marketing Research.}
    \begin{tabular}{p{0.14\linewidth} p{0.35\linewidth} p{0.45\linewidth}}
        \hline
        \textbf{Tools/Methods} & \textbf{Description} & \textbf{Studies} \\ \hline

        \multicolumn{3}{l}{\textbf{Sentiment and Emotion Analysis}} \\ \hline
        General Sentiment Analysis & Identifying and categorizing opinions expressed in a piece of text to determine the author's attitude. Widely used for understanding customer feedback and overall sentiment towards brands or products. & \cite{Nicolau2024, Park2023, Karn2023, Barbosa2023, Han2023, Solairaj2023, Khan2023, Liu2023, Galiano-Coronil2023, Rahimi2022, Sun2022, Agrawal2022, GuhaMajumder2022, Garner2022, Lamrhari2022, Alzate2022, Alantari2022, Wu2022, Zhang2022, Farzadnia2022, Celuch2021, Karthik2021, VillarroelOrdenes2021, HeidaryDahooie2021, Zhang2021, Lopez2020, Wu2020, Deng2020, Lawani2019, Kim2019, Cu2019, Ghosh2018, Peláez2018, Afolabi2017, Dubey2017, Xiang2015, Balusamy2015, Homburg2015, Yang2015} \\ 
        Aspect-Based Sentiment Analysis (ABSA) & Analyzing sentiment with respect to specific aspects or features of a product or service, providing a more granular view of customer opinions. & \cite{Liu2024, Zhai2024, Aldunate2022, Lappeman2022, Vassilikopoulou2022, Gozuacik2021, Liu2019, Chang2019, Zhang2018, Ma2017, Lee2016} \\ 
        Multimodal Emotion Recognition & Recognizing and categorizing emotions from multiple data modalities. & \cite{Chen2022} \\ \hline

        \multicolumn{3}{l}{\textbf{Topic Modeling}} \\ \hline
        Latent Dirichlet Allocation (LDA) & A generative statistical model allowing sets of observations to be explained by unobserved groups, aiding in discovering hidden topics within text data. & \cite{Le2024, Barbosa2023, Yazıcı2023, Sun2022, Swaminathan2022, Lamrhari2022, Alzate2022, Alantari2022, Wu2022, Zhang2022, Farzadnia2022, Celuch2021, Piris2021, Situmeang2020, Wu2020, Hu2019, Lee2016} \\ 
        Correlated Topic Modeling (CTM) & Analyzing the latent topical structure in text data to capture correlations among topics. & \cite{Garner2022} \\
        BERT-Based Models & Using BERT embeddings for improved semantic understanding. & \cite{Yazıcı2023} \\ 
        Top2Vec & Identifying topics using topic/word vectors. & \cite{Yazıcı2023} \\ 
        Structural Topic Modelling (STM) & Incorporating metadata to improve topic discovery and segment characterization. & \cite{Fresneda2021} \\ \hline

        \multicolumn{3}{l}{\textbf{Natural Language Generation}} \\ \hline
        
        Large Language Models (LLMs) & Using LLMs for generating marketing content and automating customer interactions. & \cite{Tafesse2024} \\
        Chatbot Integration & Developing chatbots to handle customer queries and automate responses. & \cite{Jin2023, Kushwaha2021, Chiu2021, Rhee2020} \\
        SEO Content Creation & Creating website content optimized for search engines to improve visibility and traffic. & \cite{Reisenbichler2022} \\ \hline

        \multicolumn{3}{l}{\textbf{Other Text Mining Analyses}} \\ \hline
        Text Classification & Categorizing text into predefined categories for different individualized marketing tasks. & \cite{Marandi2024, Batta2023, Loupos2023, Liu2021, Han2021, Ozan2021, Gruss2020, DeCaigny2020, Hu2019} \\ 

        Clustering & Grouping texts based on similarities. & \cite{Marandi2024, Alzate2022} \\ 
        Syntactic Surprise Analysis & Evaluating the unexpectedness of the syntactic structure within a text. & \cite{Atalay2023} \\
        Word Embedding (Word2Vec) & Representing words in vector space based on their meanings and context. & \cite{Ozan2021, Timoshenko2019} \\ 
        Word Frequency Analysis & Comparing the frequency of words in different datasets. & \cite{Cooper2022} \\ 
        Text Similarity Analysis & Measuring the similarity between texts. & \cite{Ghasemi2021, Missaoui2019} \\
        Emerging Pattern Mining & Identifying patterns that show significant changes between datasets or over time. & \cite{Li2015} \\ 

        Semantic Entity Extraction & Identifying and categorizing entities within text. & \cite{Liu2023} \\
        Text Parsing & Detect and parse attributes/entities in texts. & \cite{Cooper2022} \\ \hline

        \multicolumn{3}{l}{\textbf{Visualizations}} \\ \hline
        Word Clouds & Creating word clouds based on textual data to visualize frequently occurring terms. & \cite{Kim2022} \\ 
        Perceptual Maps & Using dimensionality reduction to plot data in 2D. & \cite{Alzate2022, Lee2016} \\ \hline
       
    \end{tabular}

    \label{tab:nlp_methods}
\end{table}

\paragraph{Other Text Mining Analyses.}

A wide array of custom text mining techniques extends beyond the already presented methods. Marketers can use advanced text classification techniques for categorizing customer feedback, detecting emerging trends, and enhancing recommender systems. This extends beyond the capabilities of conventional sentiment analysis or topic modeling. For instance, it becomes possible to create highly customized features and models to predict customers' intentions, such as the likelihood of a return visit to a hotel \citep{Marandi2024}.
In the context of unstructured data, the architectures of these models are often complex. To capture non-trivial, nonlinear, and hidden relationships within text data, deep neural networks are employed \cite{DeCaigny2020}. These models are capable of performing tasks such as sales forecasting or the construction of efficient recommender systems, as evidenced by prior research \citep{Shambour2021, Loupos2023, Missaoui2019}. 

Identifying similarities between customers, products, or brands can be performed by vector-based text similarity analysis. In order to achieve this, the respective text objects (words, sentences, collections of texts) are transformed into embeddings. There are a number of different approaches that can be employed in this regard, including the use of word embeddings with Word2Vec, document embeddings with Doc2Vec/Paragraph2Vec, and traditional lexical matching techniques such as TF-IDF \citep{Ghasemi2021}. In vectorized form, these objects can be compared and the degree of similarity quantified.
Word embeddings, for instance, can effectively capture semantic similarities between customer interactions \citep{Ozan2021, Han2021}. Additionally, using word embeddings enhances the understanding of customer needs from reviews \citep{Timoshenko2019}. Clustering algorithms can then group similar objects based on shared characteristics or behaviors, thus facilitating text-based customer and market segmentation. This enables the categorization of customers into groups based on their reviews and feedback, as well as the segmentation of brands into distinct markets \citep{Marandi2024, Alzate2022}. Furthermore, text-based similarities can be leveraged to build recommender systems based on collaborative filtering \citep{Ghasemi2021}.

Text parsing and entity extraction are also frequently used in NLP to structure information from unstructured text. This is particularly relevant for corporate reports and documents, enabling firms to automate and streamline marketing processes effectively \citep{Cooper2022, Liu2023}.

In addition to these techniques, syntactic surprise analysis and emerging pattern mining provide further capabilities for custom marketing analyses. 
The application of syntactic surprise analysis allows for the evaluation of unexpected syntactic structures within text. This can be used to improve the effectiveness of marketing messages through moderate surprise \citep{Atalay2023}. 
Emerging pattern mining focuses on significant changes in data patterns over time. With this information, marketing strategies can be proactively adapted based on shifting customer behaviors and market trends \citep{Li2015}.

\paragraph{Visualizations.}

Visualizations give marketers further guidance to effectively analyze unstructured textual data and generate actionable insights.
Techniques such as word clouds and perceptual maps help visualize frequently occurring terms and relationships within data. Word clouds are a visual representation of the most frequently occurring terms in text data and can provide an intuitive overview of customer feedback and reviews \citep{Kim2022}.

Perceptual Maps use dimensionality reduction to plot high-dimensional textual vector representations in a manner that is comprehensible to humans, typically in two dimensions. Clusters and relationships in the data can then be identified via the data points in the plot. In marketing, perceptional maps are used to illustrate customer preferences or brand positioning in a competitive market \citep{Alzate2022, Lee2016}. These maps reveal the relative positioning of products or brands, thereby highlighting areas for potential improvement or differentiation.

\subsubsection{NLP Studies by Industries}

Table \ref{table:NLP_Studies_Sector_Overview} summarizes NLP marketing studies by sector, highlighting the distribution of research efforts and sector-specific trends. Examining these studies reveals that the majority of research is not focused on banking and finance. By analyzing the approaches and findings from other sectors, we can identify valuable insights and methods that may be applicable to improving NLP-based marketing in the banking sector.

\renewcommand{\arraystretch}{1.4}
\begin{table}[ht]
\centering
\scriptsize
\caption{Overview of NLP Marketing Studies by Industry Sectors.}
\label{table:NLP_Studies_Sector_Overview}
\begin{tabularx}{\textwidth}{p{0.2\linewidth}X>{\raggedleft\arraybackslash}p{0.06\linewidth}}
\toprule

\textbf{Sector} & \textbf{Main Studies} & \textbf{Count} \\
\midrule
Automotive & \cite{Liu2024}, \cite{Sun2022}, \cite{Liu2019} & 3 \\
E-commerce & \cite{Zhai2024}, \cite{Karn2023}, \cite{Solairaj2023}, \cite{Batta2023}, \cite{Jin2023}, \cite{Liu2023}, \cite{Yazıcı2023}, \cite{Sidlauskiene2023}, \cite{Agrawal2022}, \cite{GuhaMajumder2022}, \cite{Kim2022}, \cite{Alantari2022}, \cite{Chen2022}, \cite{Karthik2021}, \cite{HeidaryDahooie2021}, \cite{Kushwaha2021}, \cite{Ghasemi2021}, \cite{Rhee2020}, \cite{Wu2020}, \cite{Kim2019}, \cite{Timoshenko2019}, \cite{Zhang2018}, \cite{Dubey2017}, \cite{Ma2017}, \cite{Yang2015} & 25 \\
Entertainment \& Events  & \cite{Loupos2023}, \cite{Celuch2021}, \cite{Deng2020} & 3 \\
Financial Services * & \cite{Li2023}, \cite{Lappeman2022}, \cite{Sajid2022}, \cite{Shumanov2022}, \cite{DeCaigny2020}, \cite{Afolabi2017} & 6 \\
Food \& Restaurants  & \cite{Han2023}, \cite{Khan2023}, \cite{Situmeang2020}, \cite{Gruss2020} & 4 \\
Hospitality \& Travel & \cite{Marandi2024}, \cite{Le2024}, \cite{Nicolau2024}, \cite{Rahimi2022}, \cite{Garner2022}, \cite{Farzadnia2022}, \cite{Vassilikopoulou2022}, \cite{Lawani2019}, \cite{Missaoui2019}, \cite{Chang2019}, \cite{Xiang2015}, \cite{Li2015} & 12 \\
Retail & \cite{Liu2021}, \cite{Chiu2021}, \cite{Han2021} & 3 \\
Technology \& Electronics & \cite{Park2023}, \cite{Zhang2022}, \cite{Gozuacik2021}, \cite{Lee2016} & 4 \\

\bottomrule
\addlinespace[0.5ex]
\multicolumn{3}{X}{* The applications in financial services will be detailed in the gap analysis in section \ref{sec:gap-analysis}.} 
\end{tabularx}

\end{table}

\paragraph{Automotive.}
The automotive sector has effectively utilized \textit{NLP} to unveil consumer preferences and analyze product defects through online reviews. Studies such as \cite{Liu2024}, \cite{Sun2022}, and \cite{Liu2019} emphasize the importance of \textit{fine-grained sentiment analysis} and \textit{consumer segmentation}. These methodologies allow automotive companies to improve product design, enhance customer relationship management, and develop targeted marketing strategies. For instance, \cite{Liu2024} introduced a framework using \textit{graph neural networks} for opinion pair generation to improve sentiment analysis accuracy. \cite{Sun2022} applied \textit{LDA}, \textit{sentiment analysis}, and \textit{cluster analysis} to segment consumers and identify key factors influencing purchasing behavior. Similarly, \cite{Liu2019} used \textit{supervised learning} and \textit{domain-specific sentiment analysis} to analyze competitive advantages from user-generated content.

\paragraph{E-commerce.}
E-commerce is the most extensively studied sector with 25 papers, demonstrating the versatile applications of NLP in this domain. The primary focus is on \textit{sentiment analysis}, \textit{recommendation systems}, and the impact of \textit{online reviews} on consumer behavior. Studies like \cite{Zhai2024} and \cite{Solairaj2023} highlight how sentiment analysis can extract consumer opinions from reviews, aiding in data-driven decisions to enhance customer satisfaction.

NLP has advanced \textit{recommendation systems}, as seen in \cite{Karn2023} and \cite{HeidaryDahooie2021}, which use sentiment analysis to improve recommendation accuracy. AI-driven \textit{chatbots} are also significant, with studies like \cite{Jin2023} and \cite{Sidlauskiene2023} exploring their role in enhancing customer interaction and personalizing shopping.

Analyzing \textit{customer feedback} is crucial for refining marketing strategies. \cite{Wu2020} and \cite{Yazıcı2023} used text-mining techniques to identify key topics and sentiments in reviews, providing insights into customer preferences and areas for improvement. Cross-cultural and \textit{multilingual} aspects are also considered, as in \cite{Le2024}, ensuring a comprehensive understanding of diverse customer bases.

Emerging technologies like \textit{voice shopping} and \textit{multimodal content analysis} are explored in studies such as \cite{Rhee2020} and \cite{Liu2023}. These innovations highlight the evolving landscape of e-commerce and the need for continuous adaptation of new technologies. Overall, the extensive research in e-commerce demonstrates NLP's significant role in enhancing online shopping by improving customer experience, personalizing marketing efforts, and driving sales through the effective use of NLP technologies.

Additionally, \cite{Timoshenko2019} demonstrates the use of NLP to identify customer needs and improve product development across various industries. By analyzing user-generated content, companies can better understand customer preferences and enhance their marketing strategies to address specific concerns and improve overall satisfaction.

\paragraph{Entertainment \& Events.}
NLP techniques are crucial in the Entertainment and Events sectors for analyzing customer reviews to enhance service quality and predict performance. Studies like \cite{Loupos2023} and \cite{Deng2020} demonstrate the impact of pre-release reviews and online word-of-mouth on box office sales. \cite{Loupos2023} highlights how critic reviews can predict movie success, while \cite{Deng2020} shows the differing impacts of user and critic reviews. In the events industry, \cite{Celuch2021} uses sentiment analysis to identify key aspects valued by event-goers, such as technical issues and customer support, aiding event organizers in improving the customer experience. 

\paragraph{Food \& Restaurant.}
NLP has been extensively applied in the food and restaurant sector to analyze customer feedback and improve service quality. Studies such as \cite{Han2023}, \cite{Khan2023}, \cite{Situmeang2020}, and \cite{Gruss2020} utilize \textit{sentiment analysis} and \textit{topic modeling} to understand customer sentiments and preferences. For instance, \cite{Han2023} examined the link between social media marketing (SMM) performance indicators and restaurant sales, finding that increased social media engagement positively correlates with sales revenue. \cite{Khan2023} used text mining to analyze online reviews from a food delivery portal, identifying key factors such as hygiene and pricing that influence delivery ratings. \cite{Situmeang2020} applied advanced text mining techniques to uncover latent dimensions of customer satisfaction from online reviews, providing insights into service improvement. \cite{Gruss2020} focused on the impact of social media post attributes on customer engagement, highlighting the importance of community-building content.

\paragraph{Hospitality \& Travel.}
The travel and hospitality sector leverages NLP to analyze customer reviews and enhance service quality. Papers like \cite{Marandi2024}, \cite{Le2024}, and \cite{Nicolau2024} use \textit{sentiment analysis}, \textit{topic modeling}, and other NLP techniques to understand customer preferences, improve service offerings, and increase customer retention. By analyzing large volumes of textual data, hotels and travel companies can better meet customer expectations and tailor their marketing strategies to different customer segments.

Studies such as \cite{Marandi2024} highlight the importance of identifying key hotel features that influence customer revisit intentions. This involves analyzing customer feedback to extract relevant features and segmenting customers based on their preferences. Similarly, \cite{Le2024} emphasizes the use of the \textit{voice of the customer} strategy to align customer expectations with service delivery, employing business intelligence solutions for detailed analysis.

Furthermore, \cite{Nicolau2024} explores the impact of daily review sentiment on hotel performance metrics, suggesting that sentiment analysis can serve as a real-time indicator of a hotel's operational effectiveness. This approach helps in dynamically adjusting pricing strategies to optimize revenue. Other studies, like \cite{Rahimi2022}, investigate gender differences in review content, providing insights into tailored marketing strategies.


\paragraph{Retail.}
The retail sector benefits from NLP through the analysis of customer engagement on social media and \textit{omni-channel marketing} strategies. Papers like \cite{Liu2021}, \cite{Chiu2021}, and \cite{Han2021} explore the impact of social media activities and customer feedback on brand engagement and sales performance. \cite{Liu2021} investigates the influence of various dimensions of a luxury brand's social media activities on customer engagement, emphasizing the importance of entertainment, interaction, and trendiness. \cite{Chiu2021} discusses the development of an \textit{omni-channel chatbot} using convolutional neural networks (CNNs) to enhance personalized service and precision marketing. \cite{Han2021} uses textual analysis to measure brand and customer focus, revealing a positive correlation between customer focus and profitability. These studies highlight the importance of understanding customer interactions and preferences to refine marketing tactics and improve customer satisfaction, which can inspire the banking sector to adopt similar approaches for enhancing customer engagement and loyalty.

\paragraph{Technology \& Electronics.}
In the technology and electronics sector, studies such as \cite{Park2023}, \cite{Zhang2022}, and \cite{Gozuacik2021} use \textit{NLP} to analyze online reviews and social media content to understand customer needs and product performance. \cite{Park2023} proposes a framework involving network analysis, \textit{topic modeling}, and \textit{sentiment analysis} to identify meaningful opinions from smart speaker reviews, helping in product improvement and brand positioning. \cite{Zhang2022} introduces a customer requirement identification framework using sentiment polarity to categorize product attributes for better product design strategies. \cite{Gozuacik2021} utilizes \textit{sentiment analysis} and opinion retrieval from social media to support product development and innovation, exemplified by the case of Google Glass. These insights help companies in product development, marketing strategy refinement, and competitive analysis, ultimately leading to better customer satisfaction and increased market share.

\vspace{1em}
The \textit{Banking and Financial Services} sector has also conducted a limited number of NLP-related studies with a focus on marketing. A more in-depth analysis combined with a gap analysis will be presented in the following section.

\section{Gap Analysis}
\label{sec:gap-analysis}

While our systematic review provided fruitful insights into the applications of analytical bank marketing and NLP across a range of industries, our investigation into the intersection of these two areas has revealed notable gaps. The literature search (\texttt{AllIntersect}, Query 3) for studies specifically addressing NLP in bank marketing revealed a lack of in-depth research with practical applications. This gap presents unique opportunities to enhance data-driven decision-making and personalize customer experiences in banking, though regulatory constraints add challenges for implementing NLP effectively in this field.

\begin{figure}[ht!]
    \centering
    \includegraphics[width=0.7\linewidth]{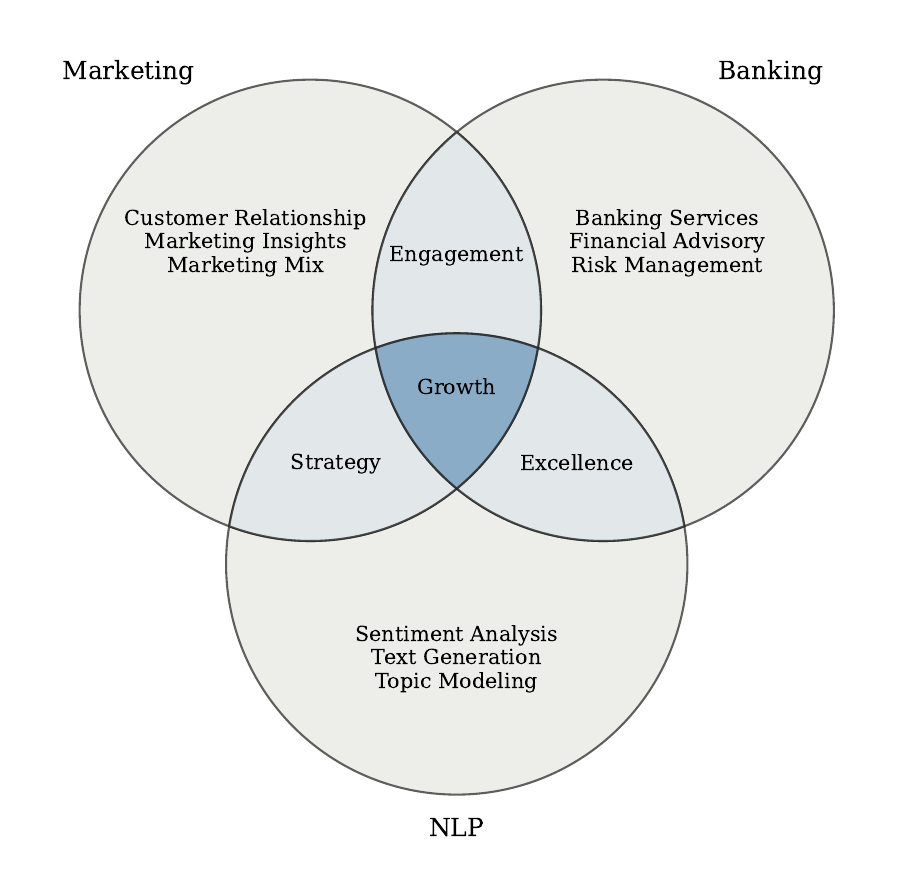} 
    \caption{The intersection of Marketing, Banking, and NLP.}
    \label{fig:venndiagram}
\end{figure}

To conceptualize the research field, Figure~\ref{fig:venndiagram} illustrates the convergence of the domains of marketing, banking, and NLP. \textit{Engagement} reflects the integration of marketing strategies to enhance the customer journey in the banking sector \citep{Kannan2017}. 
\textit{Strategy} demonstrates how NLP contributes to strategic decision-making by providing insights from textual analysis \citep{Berger2020}. \textit{Excellence} emphasizes the role of NLP in optimizing banking processes, improving efficiency, and reducing costs \citep{Kumar2016}. At the core intersection, \textit{Growth} represents the combined potential of all three domains to drive value and foster innovation in banking.
These components form the basis of our gap analysis, which explores and evaluates opportunities for NLP integration in bank marketing.

\subsection{Growth with NLP-Driven Marketing Strategies}

Integrating NLP in bank marketing stimulates growth by enhancing data-driven decision-making and personalized customer interactions. To gain a deeper insight into the potential applications of NLP within bank marketing, we conducted an AI-based mapping and visualization of the existing \texttt{MarketingBanking} and \texttt{MarketingNLP} literature. A Sentence Transformer (\textit{all-mpnet-base-v2}) was employed to process each abstract from the included studies, transforming them into high-dimensional vectors that capture semantic meanings. These vectors were then projected into a two-dimensional space using the Uniform Manifold Approximation and Projection (UMAP) technique, which preserves semantic relationships and allows for a visual representation of the research landscape.

\begin{figure}[ht!]
    \centering
    \includegraphics[width=\linewidth]{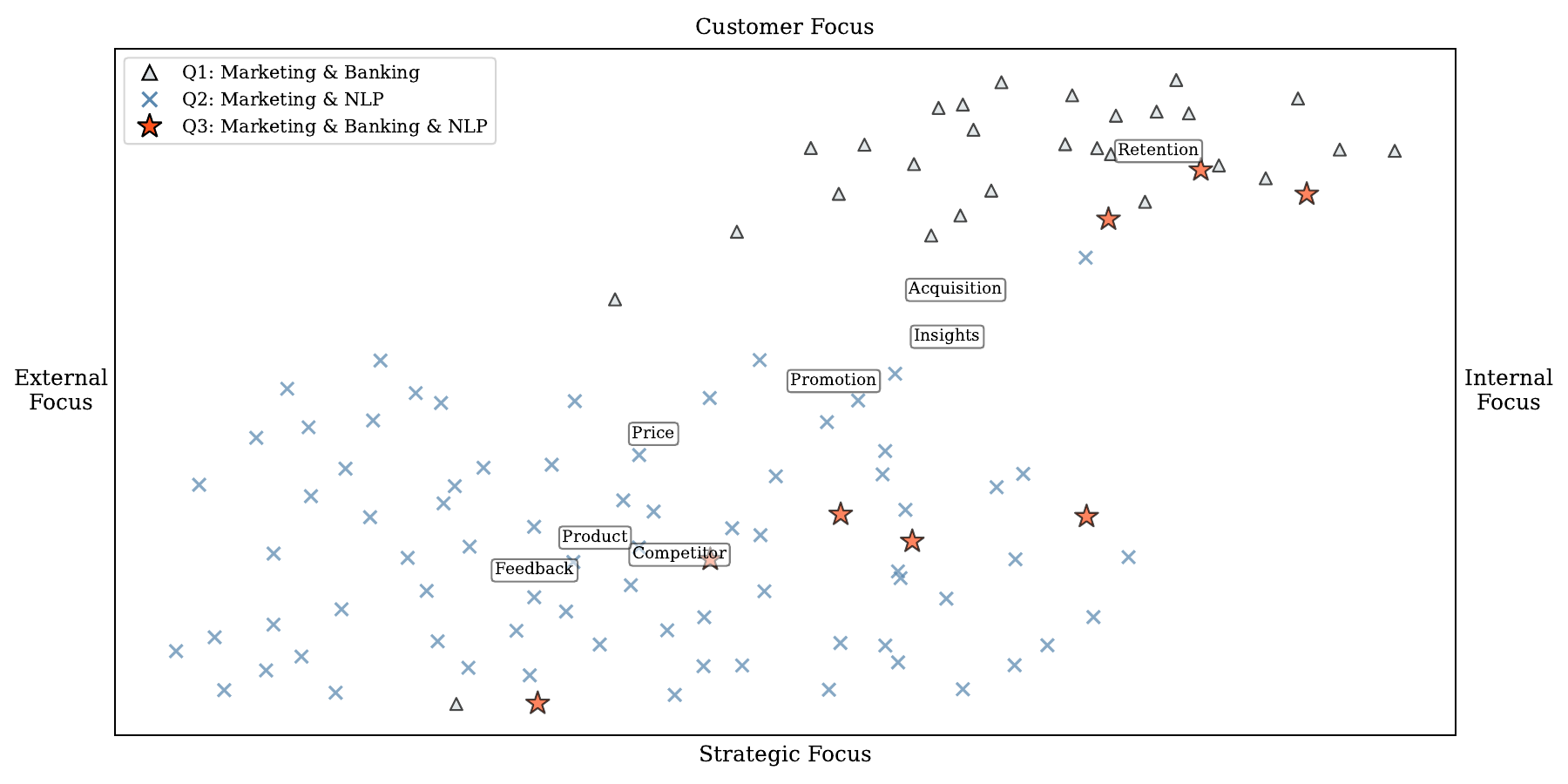} 
    \caption{Semantic Mapping of AI-driven Marketing Applications in Banking.}
    \label{fig:umap_abstracts}
\end{figure}

The resulting visualization, presented in Figure~\ref{fig:umap_abstracts}, reveals both densely covered areas, indicating well-researched topics, and sparsely populated areas that signify potential research gaps in NLP applications for bank marketing. To identify prevalent themes and research interests, we performed manual labeling on the resulting clusters. We analyzed the studies within the plot to determine relevant dimensions that best characterize the clustering patterns. Based on these analyses, key dimensions and gaps were identified to guide future studies in NLP applications for bank marketing.

The map visually positions the identified marketing categories along two strategic axes. The first axis reflects the \textit{Customer Focus vs. Strategic Focus} dimension, as supported by prior research \citep{Mustak2021}. A second axis, \textit{Internal Focus vs. External Focus}, emerges upon further analysis. The combination of these two dimensions provides a detailed understanding of the positioning of the marketing landscape, offering deeper insights into the strategic application of NLP within the banking sector.

The plot reveals clustering patterns corresponding to different marketing categories. The \textit{Analytical Marketing \& NLP} category (\texttt{MarketingNLP}) is broadly distributed across the UMAP space, showing that NLP applications in marketing are diverse and applied across a wide range of tasks. This distribution exhibits a skew towards the External Focus and Strategic Focus quadrants, indicating that NLP techniques are frequently utilized to analyze external data sources with strategic objectives in mind.

In contrast, the \textit{Analytical Marketing \& Banking} category (\texttt{MarketingBanking}) demonstrates a more concentrated clustering within the Customer Focus and Internal Focus quadrants. This reflects the traditional focus of bank marketing strategies, which are centered on customer needs and reliant on internal data sources. The overlapping studies in the \textit{Analytical Marketing \& Banking \& NLP} category (\texttt{AllIntersect}) highlight the suitability of integrating NLP into bank marketing processes, spanning both Customer Focus and Strategic Focus areas.

\begin{figure}[ht!]
    \centering
    \includegraphics[width=\linewidth]{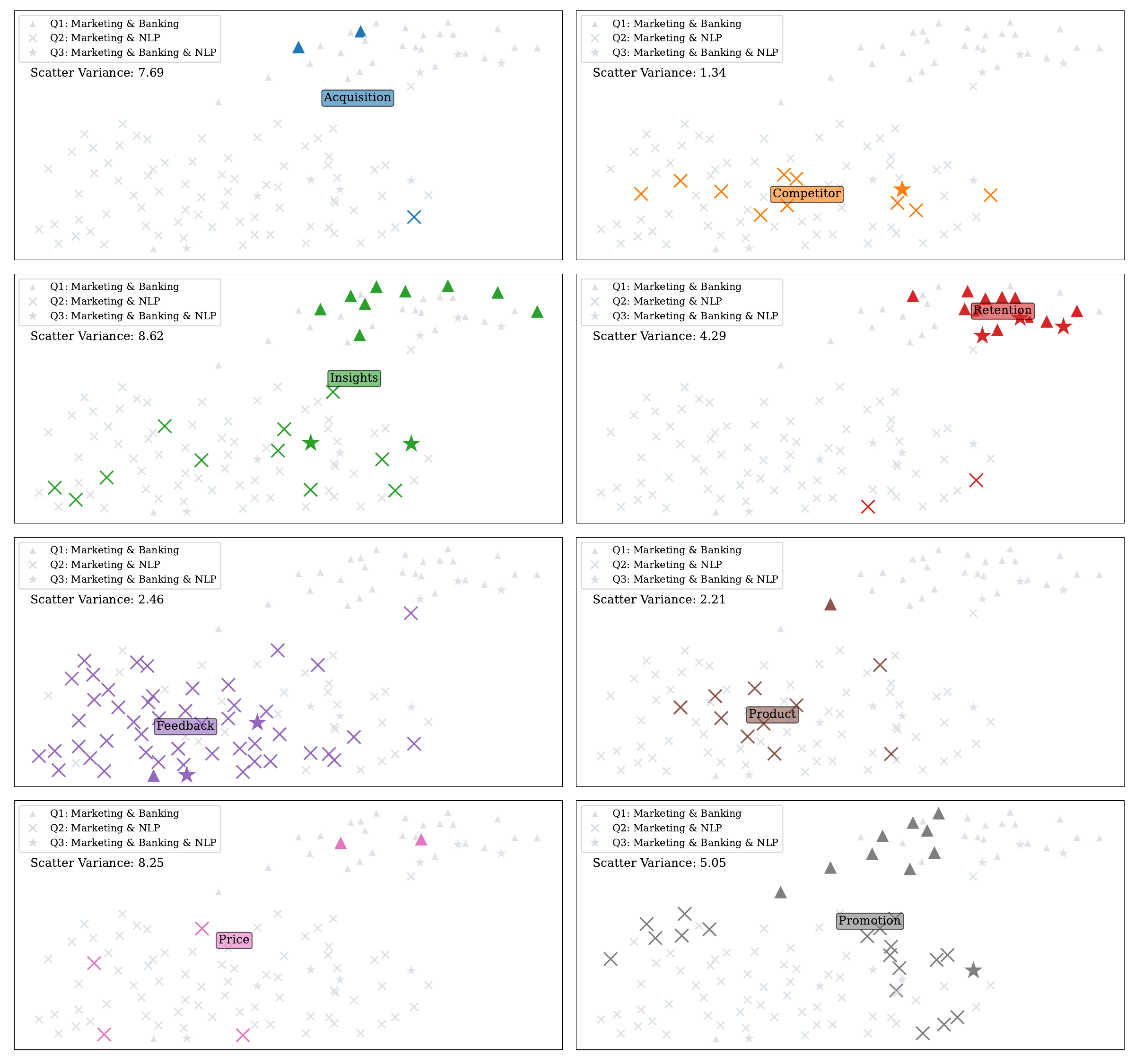} 
    \caption{Multi-Category Comparison of Semantic UMAP Embeddings.}
    \label{fig:umap_abstracts_comparison}
\end{figure}

In a more detailed examination, the UMAP projection reveals distinct patterns across multiple marketing categories, as shown in Figure \ref{fig:umap_abstracts_comparison}, which provides a category-wise comparison.

To illustrate, the \textbf{Retention} category is prominently positioned in the upper-middle section of the plot, close to the axes of Customer Focus and Internal Focus. The concentration of research and applications in this area underscores its pivotal role in banking. The \textbf{Acquisition} and \textbf{Insights} categories are located in close proximity to one another at the center of the plot, which suggests that these categories are distributed across the two dimensions in a multifaceted manner. This distribution encompasses both internal and external data, as well as a combination of direct customer engagement and strategic focus.
Similarly, the \textbf{Price} and \textbf{Promotion} categories are located at the center of the plot, reflecting different research directions and data sources. The \textbf{Product}, \textbf{Feedback}, and \textbf{Competitor} categories are more closely aligned with the strategic and external focus axis. This relationship can be attributed to the nature of feedback as an external data source, which is often used to enhance product quality. Similarly, competitor analysis requires access to external data, which is essential for informing strategic decisions.
The \textbf{Place} category has been excluded from the plot due to the assignment of only a single study to this category.

Some categories exhibit a considerable degree of variance within the plot, with studies from the same category being distributed across different quadrants. This scatter variance is also displayed in Figure \ref{fig:umap_abstracts_comparison}. It illustrates that, for instance, the Insights category, which has a high variance score, is not concentrated in any particular dimension. In contrast, the Product category demonstrates less scatter.

The UMAP plot illustrates the integration of NLP techniques in a multitude of applications in the analytical marketing landscape. However, it also reveals several gaps where further NLP adoption could enhance bank marketing. Notably, the underutilization of NLP in customer-centric applications, such as retention, acquisition, and customer insights, suggests that banks may not be fully leveraging NLP to personalize and optimize customer interactions. Addressing these gaps can enhance customer engagement and foster growth in banking, which we explore further in the following subsection.

\newpage

\subsection{Growth with NLP-Driven Customer Engagement}

To address the identified gaps in customer engagement, we now focus on the \textit{Engagement} component of our conceptual framework (Figure~\ref{fig:venndiagram}), analyzing how NLP can enhance customer interactions across the banking customer journey. Adapting the customer journey framework by \cite{Kannan2017} to a financial setting, we consider the stages of awareness, consideration, conversion, and retention. By mapping NLP applications onto these stages, we aim to provide actionable recommendations for banks to enhance engagement throughout the customer lifecycle, ultimately fostering stronger customer relationships and driving growth.

\begin{figure}[h!]
\centering
\includegraphics[trim=60 180 40 60, clip,width=1\textwidth]{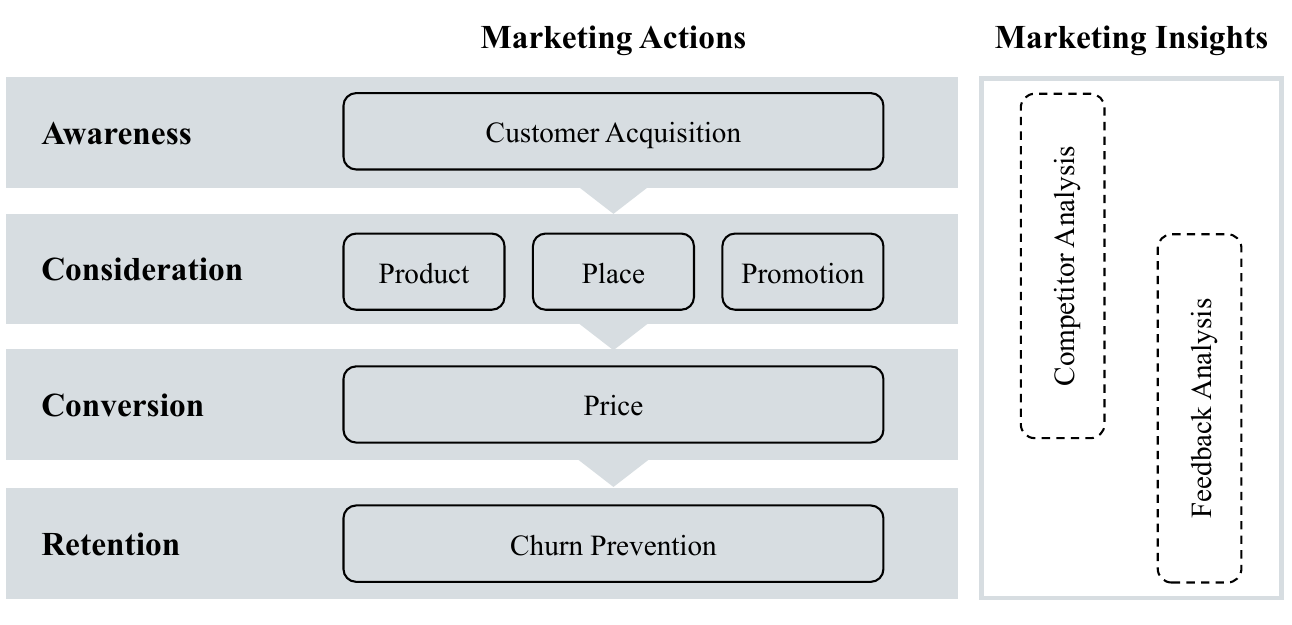}
\caption{NLP-Based Marketing Actions and Insights Across the Banking Customer Journey.}
\label{Fig:customer-journey}
\end{figure}

Figure~\ref{Fig:customer-journey} illustrates how NLP-driven marketing actions and strategic insights can enhance customer engagement at each stage of the journey.

In the \textbf{awareness} stage, banks seek to attract potential customers and establish brand recognition. Our analysis indicates that banks underutilize NLP in \textit{customer acquisition} and \textit{competitor analysis}. Traditionally, banks rely on structured data, overlooking rich insights from unstructured data sources such as social media interactions and customer reviews. This represents a missed opportunity to understand public perception and market trends.

We recommend that banks employ NLP techniques like sentiment analysis and topic modeling on social media data to gain real-time insights into customer sentiments and preferences \citep{Afolabi2017, Lappeman2022}. By identifying potential customers based on online behavior and sentiments, banks can enhance their \textit{customer acquisition} efforts. Additionally, leveraging NLP for dynamic adjustment of ad content and placement can improve engagement and acquisition rates \citep{Lamrhari2022}. For effective \textit{competitor analysis}, banks should analyze competitors' customer reviews using NLP to understand market positioning and identify areas for differentiation \citep{Schwartz2017, Quijano-Sanchez2017}.

During the \textbf{consideration} stage, potential customers evaluate banking options. Personalization and relevant information are crucial, yet banks often underutilize NLP for understanding customer needs in \textit{product} development, personalizing \textit{promotions}, and optimizing channels (\textit{place}) for delivering marketing messages.

To bridge this gap, banks should analyze unstructured customer feedback, such as online reviews and surveys, using NLP to inform \textit{product} development \citep{Zhang2022, Timoshenko2019}. Techniques like sentiment analysis and topic modeling can reveal customer preferences and areas for improvement, leading to products that better align with customer expectations \citep{HeidaryDahooie2021, Sun2022}. Furthermore, integrating NLP-driven recommender systems allows banks to offer tailored financial products and personalize \textit{promotions}, enhancing customer engagement \citep{Shumanov2022, Tafesse2024}. Optimizing \textit{place} involves using NLP to analyze customer interactions across different platforms, enabling banks to deliver promotions through the most effective channels \citep{Schwartz2017}.

At the \textbf{conversion} stage, customers decide to engage with a banking service. Here, \textit{price} is often the primary determining factor due to the commoditized nature of banking products. However, banks rarely employ NLP to analyze customer perceptions of pricing. Hence, they are missing opportunities to adjust pricing strategies based on customer feedback.

We suggest that banks apply NLP to analyze customer feedback on pricing to gain insights into perceptions of price fairness \citep{Lawani2019, Zhai2024}. By examining sentiments related to pricing, banks can adapt their \textit{price} models to align with customer expectations, thereby optimizing conversion rates. Additionally, NLP can support personalized pricing by analyzing customer interactions and behaviors, enabling a more responsive and customer-centric approach \citep{Chen2022}.

The \textbf{retention} stage is concerned with the maintenance of long-term customer relationships. Banks often use ML to detect churn signals such as reduced account activity \citep{AL-Najjar2022}, changes in spending \citep{Coşer2020}, and shifts in product usage \citep{Shirazi2019}. However, they often overlook insights from unstructured data, such as emotional cues and context.

To enhance \textit{churn prevention} and \textit{feedback analysis}, banks should integrate NLP techniques to analyze unstructured data sources like customer feedback, call center transcripts, and social media interactions. Recent studies have demonstrated improved predictive accuracy and a deeper understanding of customer behavior when textual data is integrated into the analytical process \citep{DeCaigny2020}. For instance, sentiment analysis of social media posts can reveal early signs of dissatisfaction, enabling proactive interventions \citep{Lappeman2022}. Analyzing call center logs with NLP helps identify customer concerns and allows timely corrective actions \citep{Rouhani2023}. By combining NLP with structured data models, banks can develop more advanced churn prediction systems that capture both behavioral patterns and emotional indicators \citep{Vo2021, Lamrhari2022}, ultimately enhancing customer loyalty and satisfaction.

\subsection{Growth with NLP-Driven Operational Excellence}
Operational excellence is essential for banks seeking sustainable growth and competitive advantage. By optimizing processes, reducing costs, and enhancing service quality, banks can improve financial profitability and build stronger customer relationships. While pursuing these goals, banks must also consider regulatory compliance and ethical standards as crucial preconditions, ensuring that NLP implementations are trustworthy and legally sound.

To structure our analysis, we draw on a value creation framework inspired by \cite{Kannan2017}, focusing on three dimensions: customer equity, value for the bank, and value for customers. Through our PRISMA-based literature review, we systematically identified key components within this framework, categorizing insights related to NLP applications in bank marketing. These insights align with elements of the marketing mix and highlight substantial opportunities for NLP to drive operational excellence in banking.

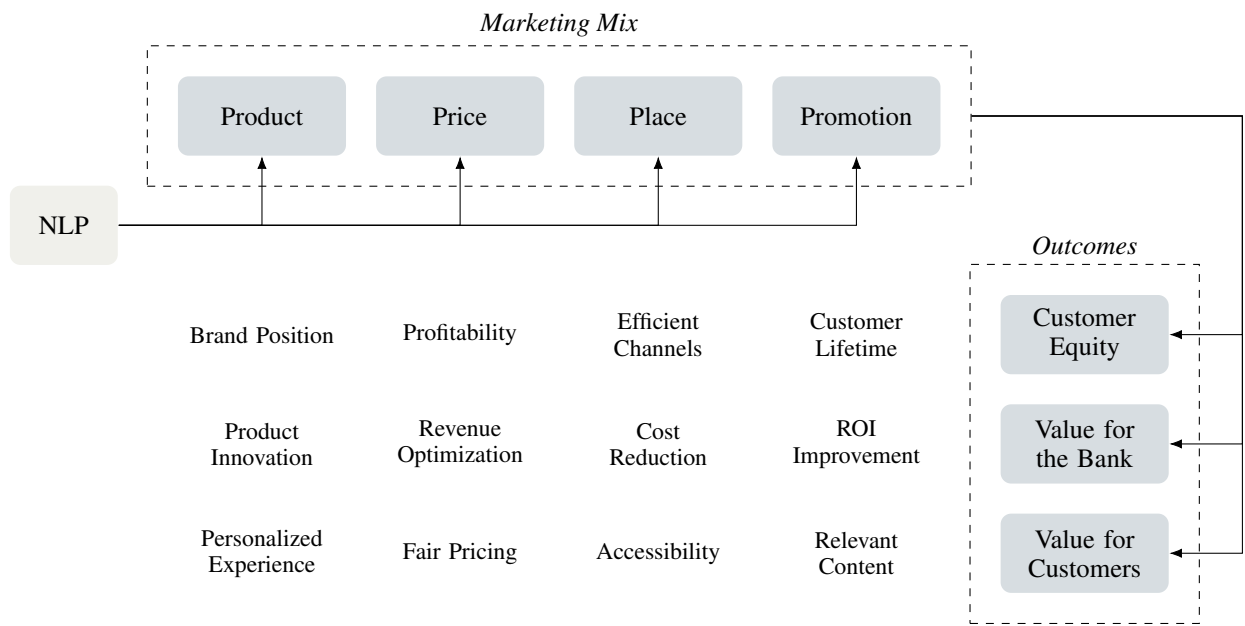
\begin{figure}[ht!]
\centering
\definecolor{hucolor2}{HTML}{d7dde1} 
\definecolor{hucolor1}{HTML}{f0f0eb} 

\begin{tikzpicture}[
    node distance=1.3cm,
    auto,
    block/.style={
        rectangle,
        align=center,
        rounded corners,
        minimum height=3em,
        text width=2cm,
        fill=hucolor2
    },
    nlpblock/.style={
        rectangle,
        thick, 
        align=center,
        rounded corners,
        minimum height=3em,
        text width=1.2cm, 
        fill=hucolor1,
    },
    nlpblock1/.style={
        rectangle,
        draw,
        align=center,
        rounded corners,
        minimum height=3em,
        text width=1cm
    },
    entry/.style={
        rectangle,
        align=center,
        minimum height=3em,
        text width=2cm,
        font=\small
    },
    line/.style={
        draw,
        -Latex
    },
    container/.style={
        rectangle,
        draw,
        dashed,
        inner sep=0.4cm
    }
]

\matrix [column sep=4mm, row sep=4mm] {
  && \node [block] (product) {Product}; & \node [block] (price) {Price}; & \node [block] (place) {Place}; & \node [block] (promotion) {Promotion}; & &\\
  
  \node [nlpblock] (nlp) {NLP}; && & & & & & \\
   && \node [entry] {Brand Position};&\node [entry] {Profitability}; &\node [entry] {Efficient\\Channels}; & \node [entry] {Customer Lifetime};& &\node [block] (customerEquity) {Customer\\Equity}; \\
  & &\node [entry] {Product\\Innovation}; &\node [entry] {Revenue \\Optimization}; &\node [entry] {Cost \\Reduction}; &\node [entry] {ROI\\ Improvement};&& \node [block] (bankValue) {Value for \\the Bank}; \\
  & &\node [entry] {Personalized \\Experience};  & \node [entry] {Fair Pricing};& \node [entry] {Accessibility};&\node [entry] {Relevant Content};&& \node [block] (valueForCustomers) {Value for\\Customers}; \\
};

\node [container, fit=(product) (price) (place) (promotion), label=above:\textit{Marketing Mix}] (marketingMix) {};

\node [container, fit=(customerEquity) (bankValue) (valueForCustomers), label=above:\textit{Outcomes}] (values) {};

\draw [line] (nlp) -| (product);
\draw [line] (nlp) -| (price);
\draw [line] (nlp) -| (place);
\draw [line] (nlp) -| (promotion);
\draw [line] ($(marketingMix.east) + (0,0)$) -- ++(3.6,0) |- (customerEquity.east);
\draw [line] (marketingMix.east) -- ++(0,0) -- ++(3.6,0) |- (bankValue.east);
\draw [line] ($(marketingMix.east) - (0,0)$) -- ++(3.6,0) |- (valueForCustomers.east);

\end{tikzpicture}

\vspace{1em}
\caption{Growth and Innovation Framework for NLP in Bank Marketing.}
\label{fig:marketing_matrix}
\end{figure}

Figure~\ref{fig:marketing_matrix} illustrates how NLP can be integrated into the marketing mix to create value across the three dimensions. The concept of \textbf{Customer Equity}, as defined by \cite{Kannan2017}, refers to the broader, long-term financial value derived from customers through effective acquisition, retention, and profitability strategies. These strategies contribute to sustained customer loyalty and higher margins over time. \textbf{Value for the Bank} refers to the economic benefits for the bank, including metrics such as sales, profits, and growth rates. \textbf{Value for Customers} is defined as the perceived benefit to the customer, encompassing improvements in financial value, brand trust, customer relationships, and overall satisfaction. This value is directly experienced through the enhanced quality of products and services offered.

\paragraph{Enhancing Customer Equity with NLP.}
We recommend that banks leverage NLP to strategically position their brand and tailor customer interactions, thereby improving marketing actions. By utilizing NLP-driven insights, banks can enhance CLV, leading to higher profitability and stronger customer relationships over time. For instance, \cite{Abdolvand2015} suggest integrating CLV as a key financial metric within CRM systems, using advanced clustering techniques to identify and retain high-value customers. Similarly, \cite{Mosavi2018} propose employing data mining techniques to create a customer value pyramid, helping banks classify and optimize the value derived from different customer segments, thereby maximizing CLV.

Moreover, banks should utilize NLP-driven insights to refine dynamic pricing strategies \citep{Ban2021}. By predicting customer behavior and aligning pricing strategies with customer expectations, banks can boost individual customer profitability and strengthen the overall equity of the customer base.

\paragraph{Driving Value for the Bank with NLP.}
To enhance financial success, banks should employ NLP to facilitate product innovation, revenue optimization, and cost reduction. By analyzing customer feedback using NLP, banks can inform product design strategies and ensure that new offerings are closely aligned with customer needs \citep{Zhang2022, Timoshenko2019}. This approach is not only crucial for successful product launches but also contributes to sustained revenue growth.

We recommend that banks adopt dynamic pricing models supported by NLP to adjust pricing in response to market conditions and customer demand, optimizing revenue streams over time \citep{Natesan2023}. Additionally, NLP offers significant cost-reduction potential, particularly through digital channels and improved customer service systems that streamline operations and reduce the need for manual intervention \citep{Hernández-Nieves2020}.

Furthermore, utilizing NLP in targeted marketing campaigns can significantly increase Return on Investment (ROI). For example, \cite{Feng2022} propose a dynamic ensemble selection method that focuses on maximizing both accuracy and profit in bank telemarketing campaigns, ensuring that marketing efforts are cost-effective and yield high returns. By optimizing resource allocation and enhancing the precision of marketing strategies, banks can achieve better financial outcomes.

\paragraph{Creating Value for Customers with NLP.}
Banks should leverage NLP to enhance the accessibility, inclusivity, and quality of banking services by providing personalized experiences tailored to individual needs. Personalized investment recommendations, driven by NLP, help customers make informed financial decisions that align with their risk profiles and financial goals \citep{Musto2015}. Additionally, integrating sentiment analysis with multi-criteria decision-making ensures that product recommendations are relevant and valuable to each customer \citep{HeidaryDahooie2021}.

Implementing fair pricing strategies is another critical area where NLP adds value for customers. By analyzing customer reviews and quality metrics, banks can ensure that pricing is transparent and aligned with customer expectations \citep{Lawani2019}, which is essential for maintaining trust and satisfaction in the banking sector.

Finally, delivering relevant content through NLP-driven strategies is crucial for customer engagement and satisfaction. Banks should use NLG to create SEO-optimized content that improves search rankings and reduces costs \citep{Reisenbichler2022}. Moreover, crafting marketing messages with an optimal level of syntactic surprise can effectively drive customer engagement \citep{Atalay2023}. Aligning chatbot personalities with user traits can further improve customer satisfaction by delivering personalized, engaging content \citep{Jin2023}. By focusing on ethical considerations, such as transparency in communication and data usage, banks can foster trust and ensure that interactions uphold high standards of fairness and integrity.

\section{Discussion and Conclusion}
\label{sec:conclusion}

AI and NLP are reshaping marketing practices across various industries. However, the banking sector has yet to fully capitalize on these technological advancements, particularly in marketing, due to unique challenges such as regulatory constraints. This study addresses a notable gap in existing research by exploring the underutilized potential of NLP in bank marketing, aiming to enhance customer engagement, boost operational efficiency, and support strategic decision-making.

Through a PRISMA-based systematic review of existing literature on AI applications in bank marketing and NLP-based marketing studies, we uncovered a noticeable scarcity of research at the intersection of these domains. Utilizing advanced semantic mapping techniques, specifically Sentence Transformers with UMAP, we identified underexplored areas where NLP can drive additional value creation in banking.

Our findings contribute to both theory and practice. Theoretically, we extend the current knowledge by connecting general marketing frameworks with banking’s unique needs. We introduced a novel growth and innovation framework for NLP integration in bank marketing, providing a conceptual model for future academic research. This framework aligns NLP applications with the marketing mix and customer journey stages, offering a structured perspective on how NLP can create value in banking.
Practically, we offer actionable insights for banking professionals. By mapping NLP applications onto the banking customer journey (awareness, consideration, conversion, and retention) we provide strategic recommendations for enhancing customer engagement and personalization. For instance, leveraging sentiment analysis and topic modeling on social media data can improve customer acquisition efforts by identifying potential customers and understanding their preferences in real-time. Additionally, integrating NLP-driven recommender systems and personalized pricing strategies can optimize conversion rates and foster stronger customer relationships.
Our analysis also highlights the potential of NLP in driving operational excellence. By optimizing processes, reducing costs, and improving service quality, NLP has the potential to enhance both profitability and customer loyalty. The proposed framework serves as a roadmap for banks to integrate NLP into their operational processes, aiming for cost reduction and revenue optimization.
In implementing these NLP-driven strategies, banks must ensure regulatory compliance and uphold ethical standards as prerequisites for maintaining customer trust and achieving long-term success.

Despite the contributions, our study has limitations that open avenues for future research. The focus on peer-reviewed articles may have excluded recent industry practices and insights presented in reports and conferences, potentially limiting the comprehensiveness of our review. Moreover, the rapid evolution of AI and NLP calls for ongoing research to keep pace with new advancements.
Future studies should explore the application of NLP in personalized marketing within the stringent regulatory constraints of the banking industry. Addressing ethical considerations, privacy issues, and compliance requirements associated with NLP-driven marketing will be essential for sustainable implementation in the industry.

\appendix



\bibliographystyle{plainnat}  
\bibliography{ref-bank-marketing, ref-nlp-marketing, ref-others, ref-reviews}

\end{document}